%% file: eccv2022submission.tex
\documentclass[runningheads]{llncs}
\usepackage{graphicx}

\usepackage{tikz}
\usepackage{comment}
\usepackage{amsmath,amssymb} 
\usepackage{color}
\usepackage{colortbl}

\usepackage{amsmath}
\usepackage{amssymb}
\usepackage{booktabs}
\usepackage{multirow}
\usepackage{threeparttable}
\usepackage{bbding}
\usepackage{makecell}
\usepackage{subcaption}
\usepackage[misc]{ifsym}
\usepackage[colorlinks,linkcolor=red]{hyperref}
\usepackage[accsupp]{axessibility}  

\newcommand{\ie}{\textit{i}.\textit{e}.}
\newcommand{\eg}{\textit{e}.\textit{g}.}
\renewcommand{\thefootnote}{}

\makeatletter
\def\blfootnote{\xdef\@thefnmark{}\@footnotetext}
\makeatother

\begin{document}
\input{0_metadata}
\input{0_abstract}
\input{1_introduction}
\input{2_related}

\input{3_method}
\input{4_results}
\input{5_conclusions}

\bibliographystyle{splncs04}
\bibliography{egbib}
\clearpage
\input{X_supplementary}
%
%
\end{document}

%% file: 0_metadata.tex
\pagestyle{headings}
\mainmatter
\def\ECCVSubNumber{5408}  
\renewcommand{\thefootnote}{}
\title{Self-slimmed Vision Transformer} 


\titlerunning{Self-slimmed Vision Transformer}
%
\author{
    Zhuofan Zong\textsuperscript{\rm 1$\star$},
    Kunchang Li\textsuperscript{\rm 3,4$\star$},
    Guanglu Song\textsuperscript{\rm 2},
    Yali Wang\textsuperscript{\rm 3,5},
    Yu Qiao\textsuperscript{\rm 3,6},
    Biao Leng\textsuperscript{\rm 1},
    Yu Liu\inst{2}\textsuperscript{\dag}
}
\authorrunning{Z. Zong, K. Li, et al.}
%
\institute{
    $^1$School of Computer Science and Engineering, Beihang University\\
    $^2$SenseTime Research\\
    $^3$ShenZhen Key Lab of Computer Vision and Pattern Recognition, SIAT-SenseTime Joint Lab,Shenzhen Institutes of Advanced Technology, Chinese Academy of Sciences\\
    $^4$University of Chinese Academy of Sciences\\
    $^5$SIAT Branch, Shenzhen Institute of Artificial Intelligence and Robotics for Society\\
    $^6$Shanghai AI Laboratory
}
\maketitle
\blfootnote{\textsuperscript{$\star$} Z. Zong and K. Li contribute equally during their internship at SenseTime.}
\blfootnote{\textsuperscript{\dag} Corresponding author.}

%% file: 0_abstract.tex
\begin{abstract}
Vision transformers (ViTs) have become the popular structures and outperformed convolutional neural networks (CNNs) on various vision tasks.
However,
such powerful transformers bring a huge computation burden,
because of the exhausting token-to-token comparison.
The previous works focus on dropping insignificant tokens to reduce the computational cost of ViTs.
But when the dropping ratio increases, 
this hard manner will inevitably discard the vital tokens, 
which limits its efficiency. 
To solve the issue,
we propose a generic self-slimmed learning approach for vanilla ViTs,
namely SiT.
Specifically,
we first design a novel Token Slimming Module (TSM),
which can boost the inference efficiency of ViTs by dynamic token aggregation.
As a general method of token hard dropping,
our TSM softly integrates redundant tokens into fewer informative ones.
It can dynamically zoom visual attention without cutting off discriminative token relations in the images,
even with a high slimming ratio.
Furthermore,
we introduce a concise Feature Recalibration Distillation (FRD) framework,
wherein we design a reverse version of TSM (RTSM) to recalibrate the unstructured token in a flexible auto-encoder manner.
Due to the similar structure between teacher and student,
our FRD can effectively leverage structure knowledge for better convergence.
Finally,
we conduct extensive experiments to evaluate our SiT.
It demonstrates that our method can speed up ViTs by $\mathbf{1.7}\times$ with negligible accuracy drop,
and even speed up ViTs by $\mathbf{3.6}\times$ while maintaining $\mathbf{97}\%$ of their performance. 
Surprisingly,
by simply arming LV-ViT with our SiT, 
we achieve new state-of-the-art performance on ImageNet.
Code is available at \url{https://github.com/Sense-X/SiT}.
\keywords{Transformer, Token Slimming, Knowledge Distillation}
\end{abstract}

%% file: 1_introduction.tex
\section{Introduction}
\label{sec:intro}

\input{compare_slim}

Since vision transformer (ViT) ~\cite{vit} started the era of transformer structure in the fundamental computer vision tasks~\cite{detr,xie2021segformer,chen2021pre}, 
variant transformers have been designed to challenge the dominance of convolutional neural networks (CNNs).
Different from CNNs that stack convolutions to encode local features progressively,
ViTs directly capture the long-term token dependencies.
However,
because of the exhausting token-to-token comparison,
current powerful transformers require huge computation,
limiting their wide application in reality \cite{levit}.
Hence,
in this paper,
we aim to design a generic learning framework for boosting the efficiency of vanilla vision transformers.

\begin{figure*}[t]
    \begin{minipage}[t]{0.28\linewidth}
        \centering
        \includegraphics[width=1\textwidth]{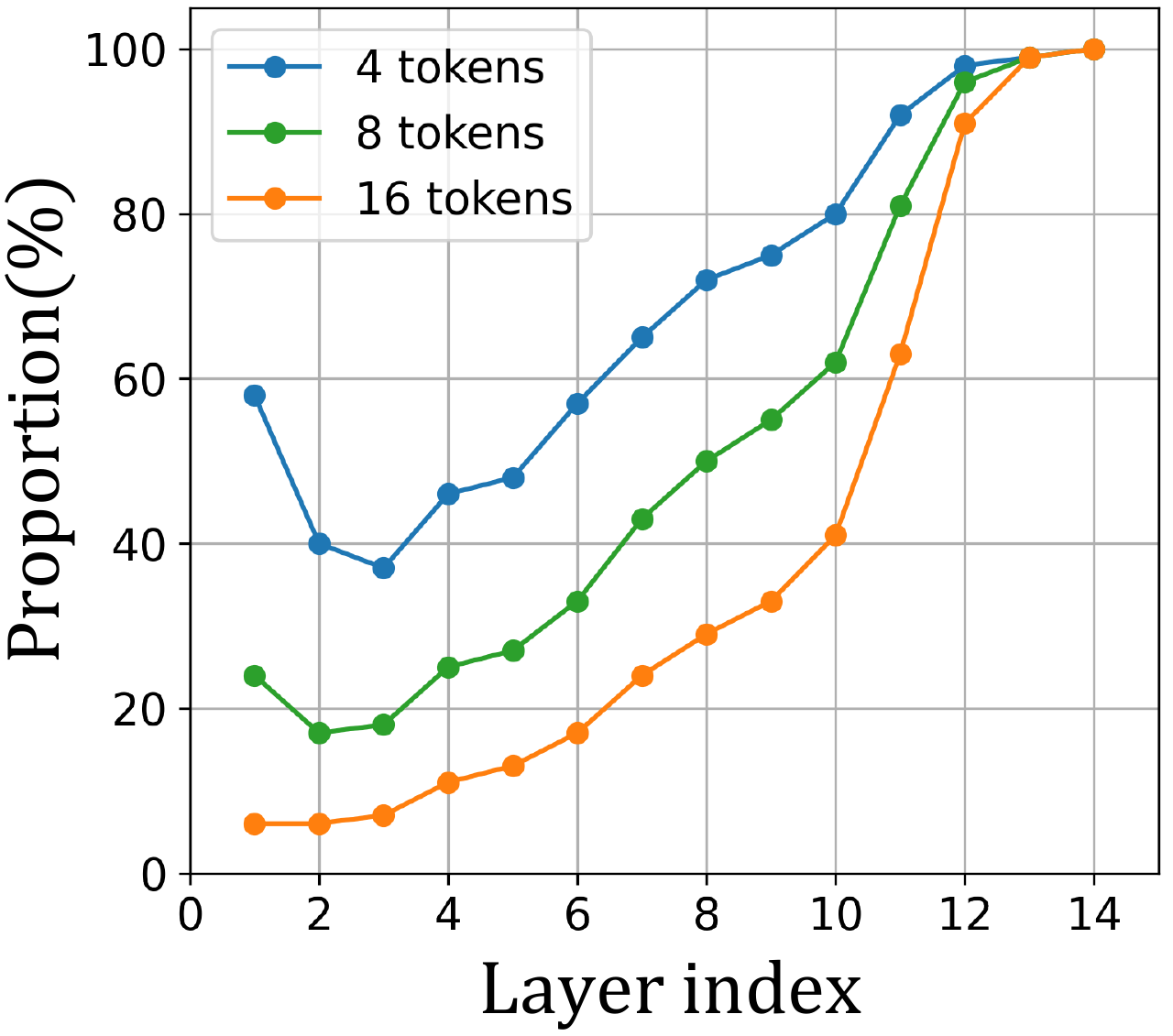}
        \subcaption{Token similarity becomes higher in deeper layers.}
        \label{similarity}
    \end{minipage}
    \begin{minipage}[t]{0.29\linewidth}
        \centering
        \includegraphics[width=1\textwidth]{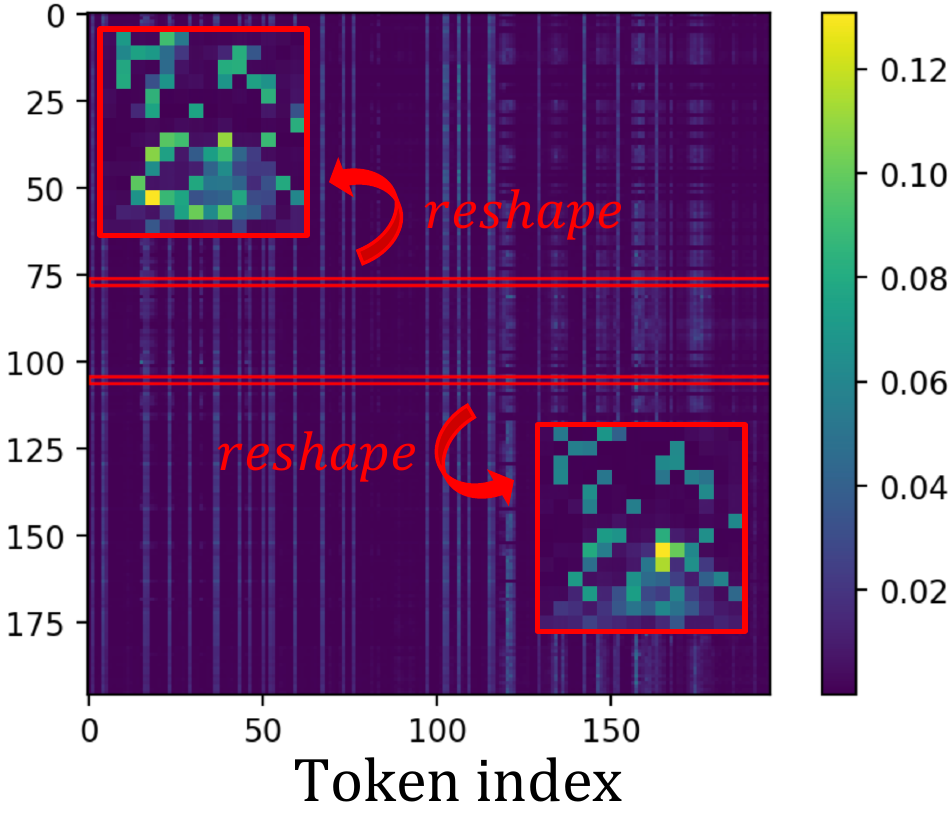}
        \subcaption{All tokens tend to focus on the same tokens in deeper layers.}
        \label{sparse_attention}
    \end{minipage}
    \begin{minipage}[t]{0.36\linewidth}
        \centering
        \includegraphics[width=1\textwidth]{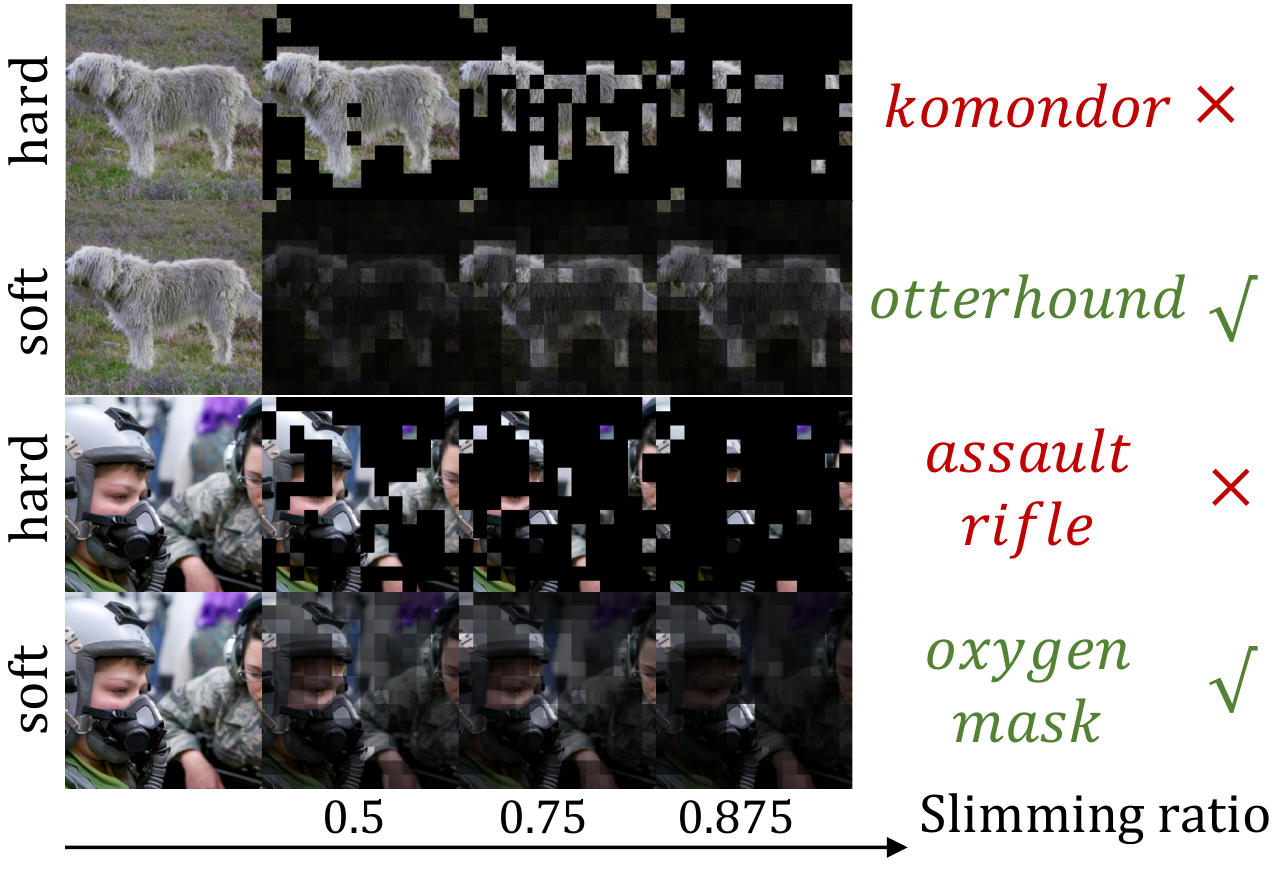}
        \subcaption{Our \textit{soft slimming} can automatically zoom the attention scope based on the object size.
        }
        \label{drop_comparison}
    \end{minipage}
    \label{fig:motivation}
    \caption{\textbf{Our motivation.}
    In Fig(a), we calculate the correlation coefficients among tokens and count the proportion that is at least similar ($\geq$0.7) to 4/8/16 tokens in different  layers.
    As for Fig(b), we randomly select two tokens in the tenth layer to show their attention.
    Moreover, we compare different token pruning methods in Fig(c). Darker tokens get less attention.
    }
\end{figure*}

To make ViTs more efficient,
we tried to explore the inherent properties of the token-to-token comparison.
We conduct a series of experiments based on LV-ViT,
which reveals that sparse attention with high token similarity exists in ViTs. 
Fig. \ref{similarity} shows that token similarity becomes higher in deeper layers,
especially in the last three layers.
Besides,
the attention tends to focus on the specific tokens in the deeper layers (Fig. \ref{sparse_attention}),
which indicates the number of decision-relevant tokens becomes fewer.
These observations demonstrate that only a few token candidates indicate meaningful information.
It inspires us a feasible structure-agnostic dimension, token number, to reduce the computational cost.
Intuitively,
we can progressively drop the redundant tokens as the network deepens.

Recent studies have tried to compress tokens via data-independent dropping with minimizing reconstruction error \cite{patchslimming},
or data-dependent dropping with differentiable scoring \cite{dynamicvit}.
However, 
data-independent dropping requires layer-by-layer optimization,
which is hard to generalize.
Moreover,
the token hard dropping will inevitably discard the vital tokens as the dropping ratio increases,
\eg,
the shape of the otterhound is destroyed in the deep layer (Fig. \ref{drop_comparison}),
thus limiting its performance as shown in Table \ref{tab:compare_slim}.

\textbf{Can we design a flexible method of token slimming,
thus decision-relevant information can be dynamically aggregated into a slimmed token set?}
To answer this question,
we propose token soft slimming.
It contains a concise Token Slimming Module (TSM),
which generates decision-relevant tokens via a data-dependent weight matrix.
As shown in Fig. \ref{drop_comparison},
by simply inserting multiple TSMs in LV-ViT,
our network can learn to localize the key object tokens.
More importantly,
the attention scope can be zoomed automatically without cutting off the discriminative token relations,
\eg,
our network can adaptively concentrate on the most informative parts of the otterhound in \textit{softer} manner,
while the oxygen mask in \textit{harder} manner.


In DynamicViT \cite{dynamicvit},
self-distillation is introduced in the last layer to minimize the performance drop brought by token sparsification.
However,
it ignores hint knowledge in the intermediate layers,
leading to inevitable knowledge loss.
To solve this issue,
we introduce a concise Feature Recalibration Distillation (FRD) to achieve stable and efficient model slimming optimization.
Note that previous hint knowledge distillation methods \cite{fitnets,at,fsp,ft} are designed for spatial structured tokens.
Since the neighbor token information is contiguous,
they can apply contiguous upsampling (\eg, deconvolution and interpolation) to find the correspondence between tokens of teacher and student.
In contrast,
our TSM generates \textit{unstructured} token set,
which can not be allocated corresponding supervision directly.
To align the token relations among unstructured tokens,
we design a reverse version of the token slimming module (RTSM) in a flexible auto-encoder manner.
Such an effective way helps our FRD densely transfer all the token information block to block.
Benefiting from the innate knowledge inheritance (structure knowledge),
our FRD is more suitable for teaching itself.
As shown in Table \ref{tab:compare_slim},
our SiT is able to improve the throughput by 43.2\% without any performance drop,
and accelerate the inference speed by over 100\% with negligible accuracy decrease.

Our self-slimmed learning method is flexible and easy to generalize to all vanilla vision transformers (SiT),
\eg,
DeiT \cite{deit},
LV-ViT \cite{lvvit} etc.
We conduct extensive experiments on ImageNet \cite{imagenet} to verify the effectiveness and efficiency.
Interestingly,
our method without self-distillation can perform even better than DynamicViT \cite{dynamicvit} with distillation.
Besides,
the SiT-XS achieves 81.8\% top-1 accuracy with $\mathbf{3.6}\times$ inference speed and SiT-L achieves competitive 85.6\% top-1 accuracy while running $\mathbf{1.7}\times$ faster.
More importantly, 
our SiT based on LV-ViT achieves the new state-of-the-art performance on ImageNet, 
surpassing recent CNNs and ViTs.

%% file: compare_slim.tex
\begin{table*}[t]
	\centering
    \caption{\textbf{Comparison to recent model pruning methods for ViT.} Our SiT surpasses all the  other methods based on structure pruning or hard dropping.}
    \setlength\tabcolsep{3pt}
    \resizebox{0.9\textwidth}{!}{
    	\begin{tabular}{l|l|c|c|cc}
            \Xhline{0.8pt}
    		\multirow{2}*{Type} & \multirow{2}*{Method} & \multirow{2}*{Reference} &  ImageNet & \multicolumn{2}{c}{Throughput} \\
    		~ & ~ & ~ & Top-1(\%) & (image/s) & (\%) \\
    		\hline
    		Baseline & DeiT\cite{deit} & ICML21 & 79.8 & 1637 & 0 \\
    		\hline
    		Structure Pruning & S$^2$ViTE\cite{chasing} &  NeurIPS21 & 79.2\color[RGB]{192, 57, 43}{(\textbf{$-$0.6})} & 2117 & 29.3 \\
    		\hline
    		\multirow{5}*{Token Hard Dropping} & PS-ViT\cite{patchslimming} &  CVPR22 & 79.4\color[RGB]{192, 57, 43}{(\textbf{$-$0.5})} & 2351 & 43.6 \\
    		~ & IA-RED$^2$\cite{ia_rea} & NeurIPS21 & 79.1\color[RGB]{192, 57, 43}{(\textbf{$-$0.7})} & 2369 & 44.7 \\
    		~ & Dynamic-ViT\cite{dynamicvit} & NeurIPS21 & 79.3\color[RGB]{192, 57, 43}{(\textbf{$-$0.5})} & 2575 & 57.3 \\
    		~ & Evo-ViT\cite{evovit} & AAAI22 & 79.4\color[RGB]{192, 57, 43}{(\textbf{$-$0.4})} & 2629 & 60.6 \\
    		~ & EViT\cite{evit} & ICLR22 & 79.1\color[RGB]{192, 57, 43}{(\textbf{$-$0.7})} & 2783 & 70.0 \\
    		\hline
    		\multirow{2}*{Token Soft Slimming} & \multirow{2}*{Our SiT} & ECCV22 &  79.8\color[RGB]{17, 122, 101}{(\textbf{$-$0.0})} & 2344 & 43.2 \\
    		~ & ~ & ECCV22 &79.4\color[RGB]{192, 57, 43}{(\textbf{$-$0.4})} & 3308 & \color[RGB]{17, 122, 101}{\textbf{102.1}}  \\
            \Xhline{0.8pt}
    	\end{tabular}
    }
    \label{tab:compare_slim}
    \vspace{-0.3cm}
\end{table*}

%% file: 2_related.tex
\section{Related Works}
\label{sec:related}

\subsection{Vision Transformers}
Transformer architecture \cite{attention} was first proposed for machine translation. 
The success of transformer in NLP inspires the application of transformers in various vision tasks,
for example, 
DETR \cite{detr} for object detection and ViT \cite{vit} for image recognition.
ViT is the first pure transformer that achieves the state-of-the-art performance on ImageNet \cite{imagenet}.
Recent ViT variants mainly focus on better optimization and more powerful performance \cite{deepvit,volo,t2t,pvt,tnt,crossvit,cswin,cvt,convit,twins,focal,pit,localvit,uniformer}. 
However, 
few of them explore to improve the efficiency of vision transformers \cite{levit}. 
In this paper, 
we aim to design a general optimization framework named self-slimmed learning to promote the efficiency of ViTs.

\begin{figure*}[tp]
    \centering
    \includegraphics[width=0.9\textwidth]{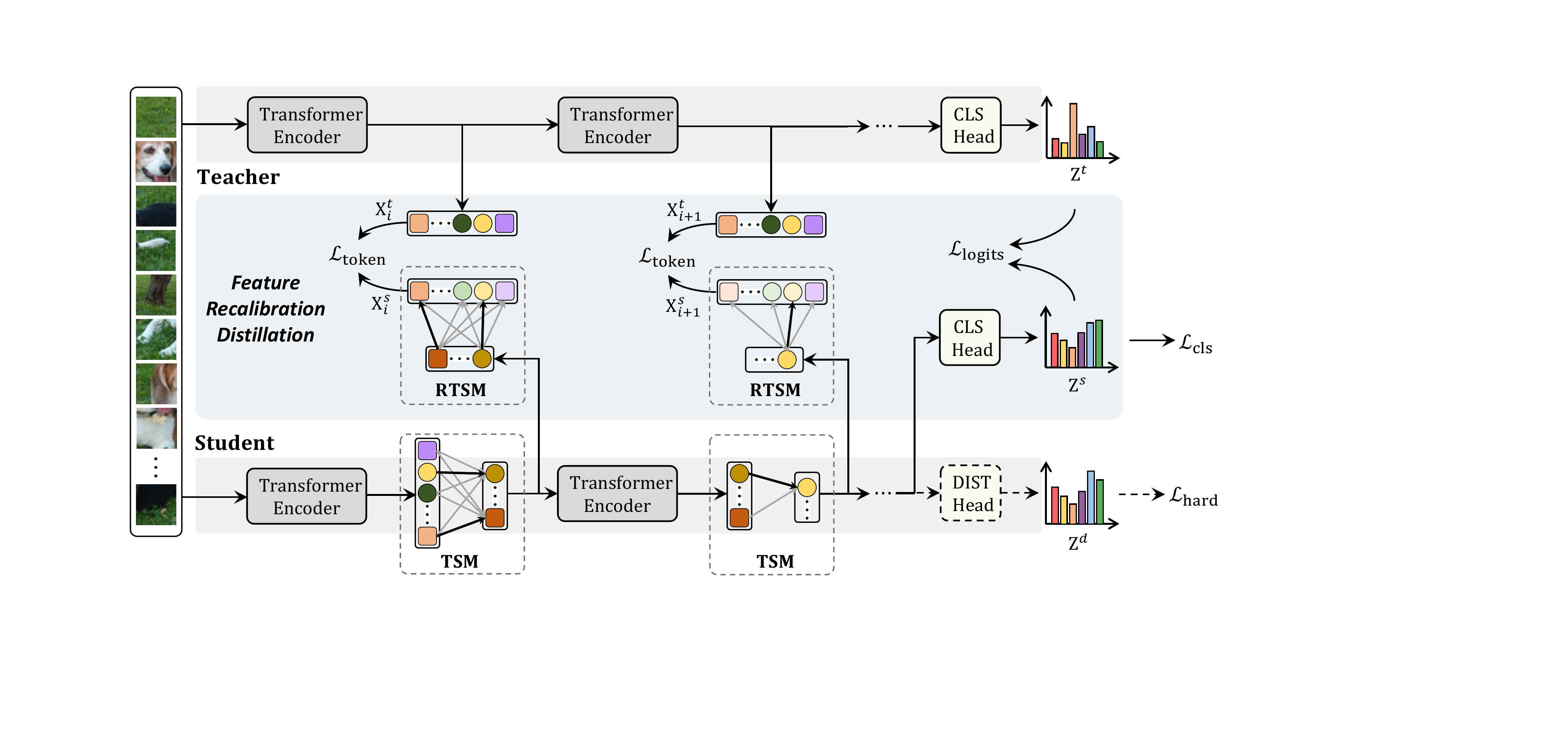}
    \caption{\textbf{The framework of our self-slimmed learning.} We insert our token slimming modules (TSM) into vanilla vision transformer. 
    To reduce decision-relevant information loss,
    we apply Feature Recalibration Distillation (FRD),
    wherein the reverse version of TSM (RTSM) is utilized to recalibrate unstructured token.
    The dash lines indicate the prediction supervision from an extra CNN teacher is optional and complementary to our method.}
    \label{fig:framework}
\end{figure*}

\subsection{Transformer Slimming}
The large computation of self-attention hinders the wide application of ViTs,
such as detection and segmentation with the high-resolution input image.
To solve this problem, 
several prior works concentrate on designing sparse attention \cite{knn_sa,swin} or structure pruning \cite{chasing}. 
S$^2$ViTE \cite{chasing} dynamically extracts and trains sparse subnetworks of ViTs, while sticking to a fixed small parameter budget. 
However, model structure pruning struggles to trim down the inference latency.
Other works try to reduce the token redundancy \cite{dynamicvit,patchslimming,ia_rea,evovit} by entirely dropping the unimportant tokens,
which brings more improvements on throughput compared to structure pruning.
Different from the above works,
our SiT aggregates all tokens into fewer informative tokens in a soft manner by a concise slimming module.
It can automatically zoom the attention scope to localize the key object for better recognition.

%% file: 3_method.tex
\section{Method}
In this section, we describe our self-slimmed learning for vision transformer (SiT) in detail. First, we introduce the overall architecture of SiT. Then, we explain the vital design of our SiT, \ie, token slimming module (TSM) and feature recalibration distillation (FRD). Finally, we thoroughly compare our TSM and FRD with other counterparts.

\subsection{Overview of Self-slimmed Learning}
The overall framework is illustrated in Fig. \ref{fig:framework}. 
We first design a lightweight Token Slimming Module (TSM) for conventional ViTs to perform token slimming,
and its reverse version (RTSM) to recalibrate unstructured tokens for hint knowledge distillation. 
We divide the slimmed vision transformer into multiple stages (\eg, 4 stages as in prior works \cite{levit,swin}), 
where different numbers of tokens participate in feed-forward computation.
To decrease the information loss,
we propose a
block-to-block 
feature recalibration distillation (FRD),
wherein the original vision transformer can serve as a teacher to minimize the difference between itself and the slimmed student.
Finally, 
we integrate TSM and FRD to form a general self-slimmed learning method for all vanilla ViTs. 

\subsection{Token Slimming Module}

Given a sequence of $N$ input tokens with $C$ channels
$\mathbf{X}=[\mathbf{x}_{1}; \mathbf{x}_{2}; \cdots; \mathbf{x}_{N}] \in \mathbb{R}^{N\times C}$,
(class token is omitted as it will never be pruned),
token slimming aims to dynamically aggregate the redundant tokens to generate $\hat{N}$ informative tokens 
$\hat{\mathbf{X}}=[\hat{\mathbf{x}}_{1}; \hat{\mathbf{x}}_{2}; \cdots; \hat{\mathbf{x}}_{\hat{N}}]$:
\begin{equation}
  \hat{\mathbf{X}} = \hat{\mathbf{A}}\mathbf{X},
\end{equation}
where $\hat{\mathbf{A}} \in \mathbb{R}^{\hat{N} \times N}$ is a normalized weight matrix: 
\begin{equation}
    \sum_{i=1}^{\hat{N}}\hat{\mathbf{A}}_{i,j}=1,\quad \texttt{where}\ \ j=1 \dots N.
\end{equation}

\begin{figure*}[tp]
    \centering
    \includegraphics[width=0.9\textwidth]{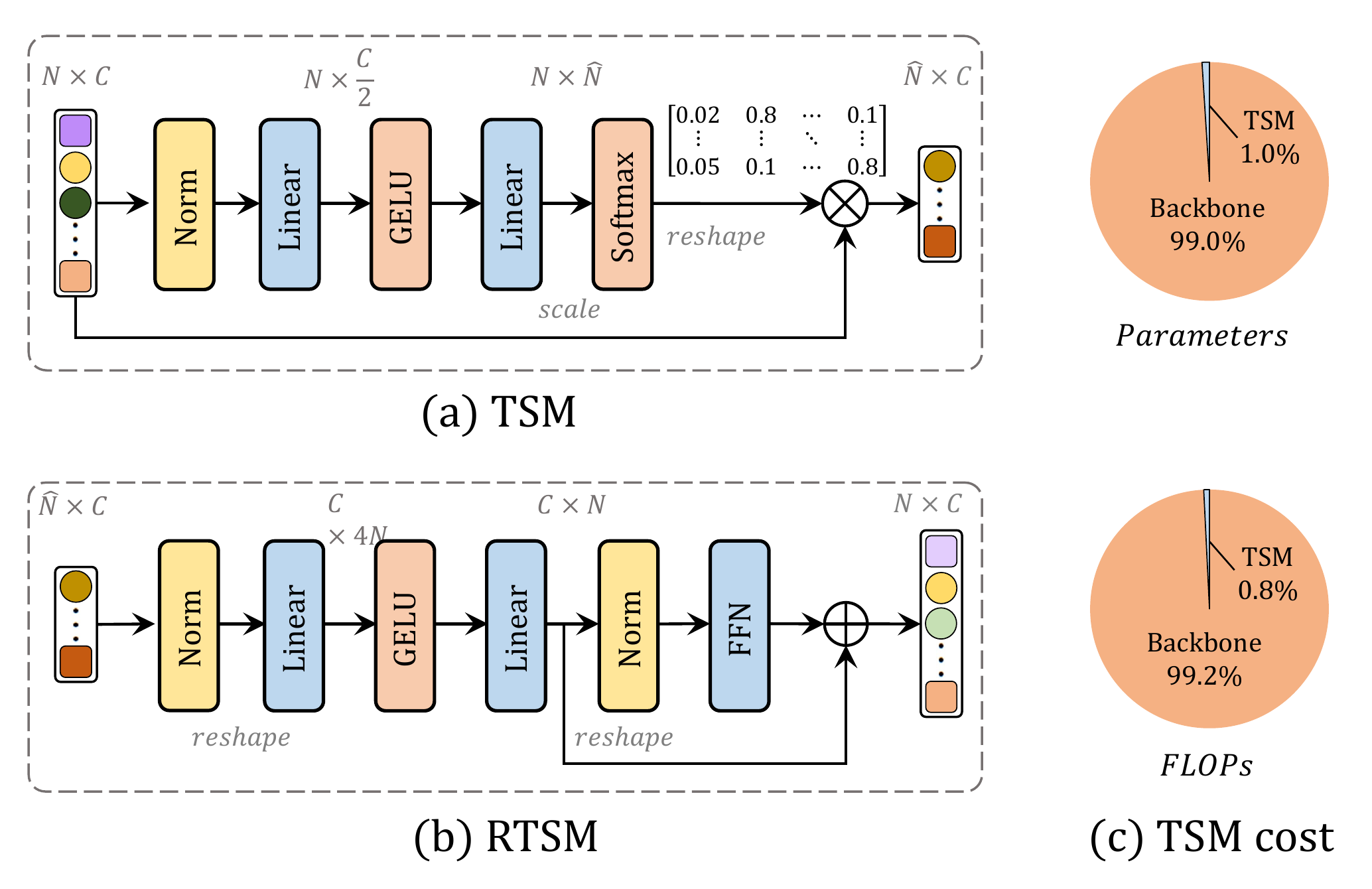}
    \caption{The token slimming module (TSM) and its reverse version (RTSM).}
    \label{fig:module}
\end{figure*}

Such operation is differentiable and friendly to end-to-end training.
We follow the design paradigm of self-attention \cite{transformer} and propose a lightweight token slimming module (TSM) shown in Fig. \ref{fig:module}{\color{red}a}:
\begin{equation}
    \hat{\mathbf{A}} = {\rm Softmax}(\frac{W_{q}\sigma (\mathbf{X}W_{k})^{T}}{\tau}),
\end{equation}
where $W_{k} \in \mathbb{R}^{C \times \frac{C}{2}}$ and $W_{q} \in \mathbb{R}^{\hat{N} \times \frac{C}{2}}$ are both learnable parameters. $\sigma$ and $\tau$ represents the nonlinear function (GELU) and scaling factor respectively. 
Similar to self-attention,
TSM generates a global attention matrix, 
but it requires much fewer overhead in terms of throughput and memory usage during both training and inference.
Fig. \ref{fig:module}{\color{red}c} shows that TSM blocks only require negligible cost.
Thanks to the learnable scaling factor $\tau$,
the attention tends to be sparse in our experiments,
which means it learns to focus on the most informative tokens.

\noindent
\textbf{Hard dropping vs. soft slimming.}
The prior works have tried to compress tokens via hard dropping \cite{patchslimming,dynamicvit},
in which the slimming weight $\hat{\mathbf{A}}_{i,j}\in\{0, 1\}$ is a binary decision matrix,
\ie,
dropping or keeping the corresponding token. 
However, 
this approach with binary decision leads to severe information loss if numerous tokens are discarded.
Such weakness limits their high efficiency on ImageNet \cite{imagenet},
wherein the objects often occupy a large part in the pictures.
On the contrary,
we design soft slimming with a learnable normalized weight $\hat{\mathbf{A}}_{i,j} \in (0, 1)$,
which is able to discriminate the meaningful tokens in a global view.
As shown in Fig. \ref{drop_comparison},
our soft slimming can dynamically zoom the attention scope to cover the significant regions for classification.
It reveals that $\hat{\mathbf{A}}$ can adaptively become a one-hot matrix to help our SiT focus on the most informative part.

\subsection{Feature Recalibration Distillation}
\textbf{Feature recalibration.}
Though token slimming significantly reduces the inference latency,
when using a large slimming rate,
it inevitably decreases the accuracy because of the information loss.
Hint knowledge distillation is the common method to maintain meaningful information in intermediate layers,
wherein the challenge is to align the feature semantics between student and teacher.
Previous works \cite{fitnets,fsp} adopt spatial deconvolution or interpolation to cope with this misalignment between spatial contiguous features.
However, 
it is not suitable for slimmed unstructured tokens with spatially discrete semantics.
To solve this problem,
we design a reverse version of the token slimming module (RTSM) to recalibrate the unstructured tokens in a flexible auto-encoder manner (Fig. \ref{fig:module}{\color{red}b}).
Therefore, 
all the token information can be seamlessly transferred from the teacher. 
Note that we only perform RTSM when training,
thus no extra computation is introduced during inference.
We first linearly transform the informative tokens into plenty of token candidates,
thus utilizing a non-linear function (GELU) to filter the vital representations.
Finally,
another linear transformation is performed to compress the token candidates:
\begin{flalign}
  \hat{\mathbf{X}}' = \mathbf{A}_2(\sigma(\mathbf{A}_1\hat{\mathbf{X}})),
\end{flalign}
where $\mathbf{A}_1 \in \mathbb{R}^{4N \times \hat{N}}$ 
and $\mathbf{A}_2 \in \mathbb{R}^{N \times 4N}$ in our experiments.
To further enhance the token representations,
we introduce an extra multi-layer perception (MLP) block \cite{transformer} with residual connection \cite{resnet}:
\begin{flalign}
  \mathbf{X}' = \hat{\mathbf{X}}'+{\rm MLP}(\hat{\mathbf{X}}'). \label{recover_token}
\end{flalign}
The recalibrated features $\mathbf{X}'$ will be forced to be consistent with the teacher features in FRD,
ensuring the sufficient information of the slimmed tokens $\hat{\mathbf{X}}$.

\noindent
\textbf{Knowledge distillation.} 
Due to the invariant model structure, 
we design a block-to-block knowledge distillation for the recalibrated features:
\begin{equation}
    \mathcal{L}_{\rm token} = \frac{1}{LN}\sum_{i=1}^L
    \sum_{j=1}^N(\mathbf{X}^{s}_{i,j}-\mathbf{X}^{t}_{i,j})^{2},
\end{equation}
where $\mathbf{X}^{s}_{i,j}$ and $\mathbf{X}^{t}_{i,j}$ refer to the $j$-th token embedding at the $i$-th layer of the student and teacher, respectively.
$L$ means the layer number.
Note that $\mathbf{X}^{s}$ refers to the recalibrated tokens $\mathbf{X}'$ in Eq. \ref{recover_token}.
With such reconstruction loss,
the student model will be forced to maintain as much as knowledge in the informative tokens $\hat{\mathbf{X}}$. 
Besides, to further alleviate the classification performance deterioration caused by token slimming, 
we introduce the logits distillation to minimize the predictions difference between the student and teacher:
\begin{equation}
    \mathcal{L}_{\rm logits} = {\rm KL}(\psi(\mathrm{Z}^{s}), \psi(\mathrm{Z}^{t})),
\end{equation}
where KL denotes Kullback–Leibler divergence loss and $\psi$ is the softmax function. 
$\mathrm{Z}^{s}$ and $\mathrm{Z}^{t}$ are respectively the predictions of the student and teacher model.
Moreover,
the above FRD is complementary to the hard distillation in \cite{deit}:
\begin{equation}
    \mathcal{L}_{\rm hard} = {\rm CrossEntropy}(\psi(\mathrm{Z}^{d}),y^{c}),
\end{equation}
where $\mathrm{Z}^{d}$ indicates the prediction of distillation head and $y^{c}$ is a hard decision of the extra CNN teacher.
It can further improve the performance with longer training epochs.
Our final objective of distillation for self-slimmed learning is:
\begin{equation}
    \mathcal{L}_{\rm dist} = 
    \lambda_{\rm token} \mathcal{L}_{\rm token}+
    \lambda_{\rm logits} \mathcal{L}_{\rm logits}+
    \lambda_{\rm hard} \mathcal{L}_{\rm hard},
\end{equation}
where $\lambda$ is the coefficient balancing the three distillation losses. 
We set $\lambda_{\rm logits}=2,\lambda_{\rm token}=2$ by default.
$\lambda_{\rm hard}$ is set to 1 when the CNN teacher is involved.
As for the training objective of self-slimmed learning, we treat the classification task and the distillation task equally:
\begin{flalign}
    \mathcal{L}_{\rm cls} 
    &= {\rm CrossEntropy}(\psi(\mathrm{Z}^{s}),\overline{y}), \\
    \mathcal{L}_{\rm global} 
    &= \mathcal{L}_{\rm cls}+\mathcal{L}_{\rm dist},
\end{flalign}
where $\overline{y}$ means the ground truth, \ie, one-hot label.

\begin{figure*}[tp]
    \centering
    \begin{minipage}[t]{0.46\linewidth}
        \centering
        \includegraphics[width=1\textwidth]{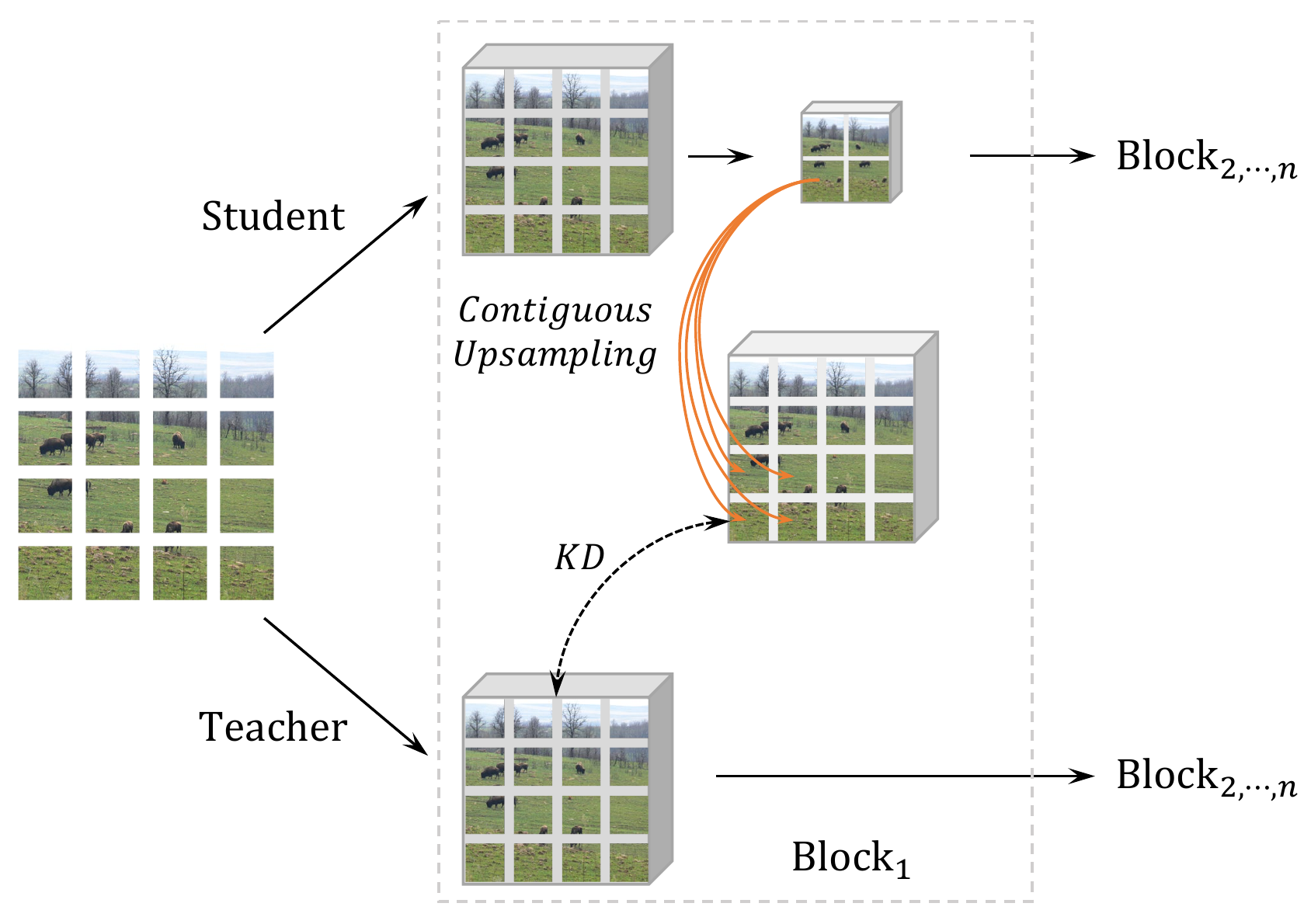}
        \subcaption{Structured Tokens.}
        \label{structured_tokens}
    \end{minipage}
    \begin{minipage}[t]{0.46\linewidth}
        \centering
        \includegraphics[width=1\textwidth]{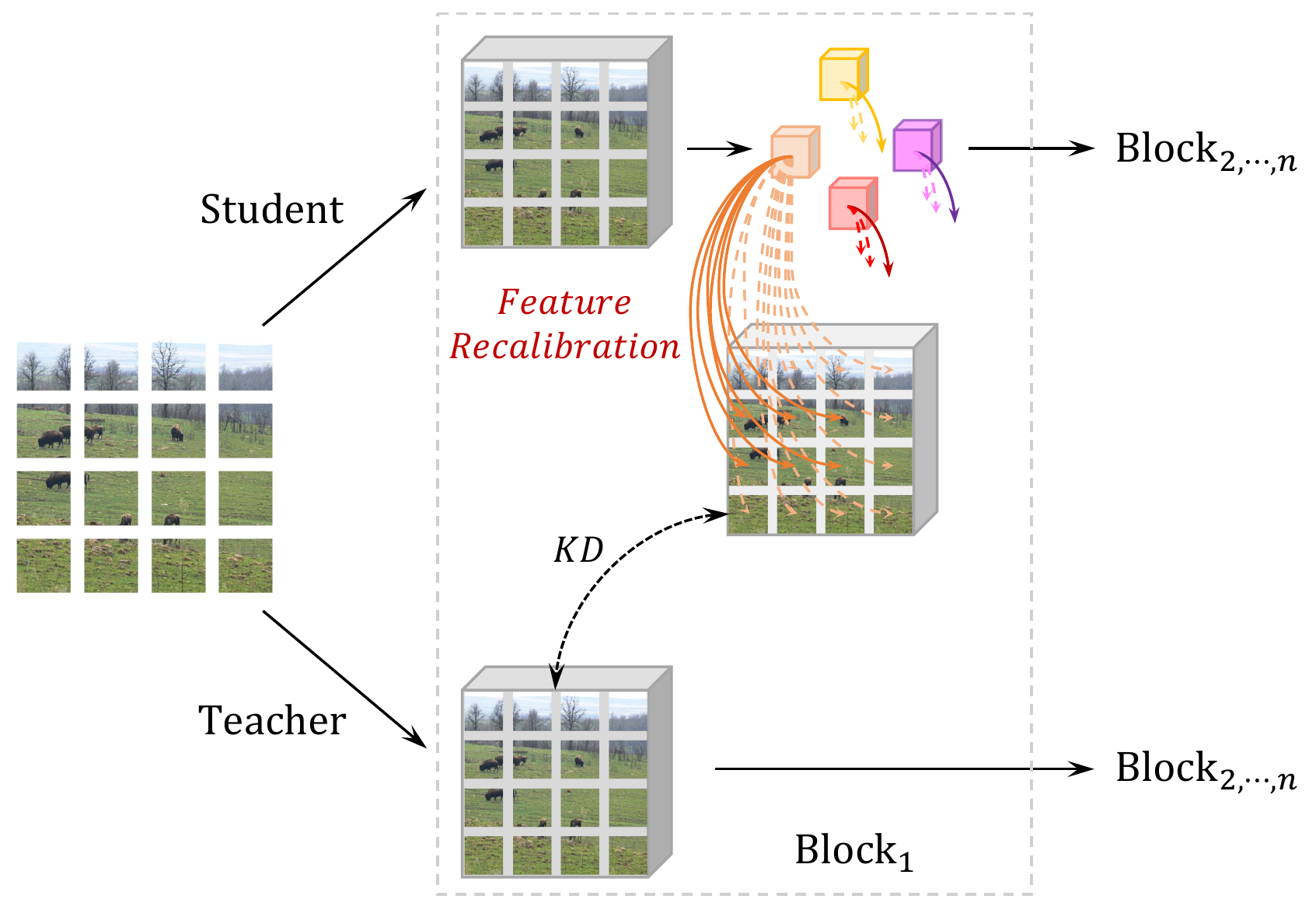}
        \subcaption{Unstructured Tokens.}
        \label{unstrucutred_token}
    \end{minipage}
    \caption{Hint knowledge distillation for structured and unstructured tokens.}
\end{figure*}

\noindent
\textbf{FRD vs. other knowledge distillation.}
Firstly,
current vision transformers \cite{deit,cait} simply select a strong teacher network with totally different architectures,
such as RegNet for DeiT.
Only a few knowledge (\eg, final predictions) can be inherited,
thus the semantic information in the intermediate layer is ignored.
In FRD, 
thanks to the consistency between the teacher and student,
we naturally conduct densely token-level supervision for each block,
which greatly improves the stability of the model mimicking.
Besides,
the popular hint knowledge distillation method \cite{fitnets,fsp} are mainly designed for spatial structured tokens.
As shown in Fig. \ref{structured_tokens},
they can simply apply local and contiguous upsampling to reconstruct tokens.
However,
as shown in Fig. \ref{unstrucutred_token},
the token slimming generates an unstructured token set.
Each token contains partial information of previous tokens.
To recalibrate the unstructured features,
we design a concise RTSM in a flexible auto-encoder manner.
Thus via reconstruction loss,
our FRD can force the student model to maintain sufficient knowledge in the informative tokens.

%% file: 4_results.tex
\section{Experiments}

\input{sota_imagenet}

\subsection{Implementation Details}
In this section, 
we conduct comprehensive experiments to empirically analyze the effectiveness of our proposed self-slimmed learning for vision transformer (SiT). 
All the models are evaluated on the ImageNet dataset \cite{imagenet}. 
For our teacher models,
we train LV-ViTs \cite{lvvit} following the original settings,
but we replace the patch embedding module with lightweight stacked convolutions inspired by LeViT \cite{levit}. 
As for student models,
all the training hyper-parameters are the same as DeiT \cite{deit} by defaults. 
For initialization,
we load all the weights from the corresponding teacher models to accelerate the convergence and train them for 125 epochs.
If utilizing an extra CNN teacher,
we extend the training time to 300 epochs for better improvements.
Moreover,
we set different initial learning rates for the backbone and the feature recalibration branch,
which are $0.0002 \times \frac{batch\ size}{1024}$ and $0.001 \times \frac{batch\ size}{1024}$ respectively. 
For token slimming, 
we insert TSM three times,
thus there are four stages in SiT.
The default keeping ratio $\hat{N}/N$ is set to 0.5,
which means the token number is halved after slimming.

\subsection{Main Results}
We conduct our self-slimmed learning for LV-ViT \cite{lvvit},
which is the state-of-the-art vanilla vision transformer.
Table \ref{main_results} shows our detailed settings for different SiT variants.
For SiT-Ti and SiT-XS,
we explore their capacity for fast inference,
thus we insert TSMs in the early layers.
It demonstrates that our self-slimmed method is able to speed up the original vision transformers by $\mathbf{3.6}\times$,
while maintaining at least 97\% of their accuracy.
Besides,
we adopt another CNN teacher to provide the hard label as in DeiT \cite{deit}.
The results show that complementary prediction supervision can further improve performance.
As for other variants,
we insert TSMs in the deeper layers.
Surprisingly,
with negligible accuracy drop,
our SiTs are up to $\mathbf{1.7}\times$ faster than their teacher models.
It is worth mentioning that,
extra CNN distillation brings little improvement,
mainly because the CNN teacher is inferior to the original transformer teacher ($82.9\%$ \textit{vs.} $83.3\%/84.2\%$).

\begin{figure*}[tp]
        \centering
    \begin{minipage}[t]{0.44\linewidth}
        \centering
        \includegraphics[width=1\textwidth]{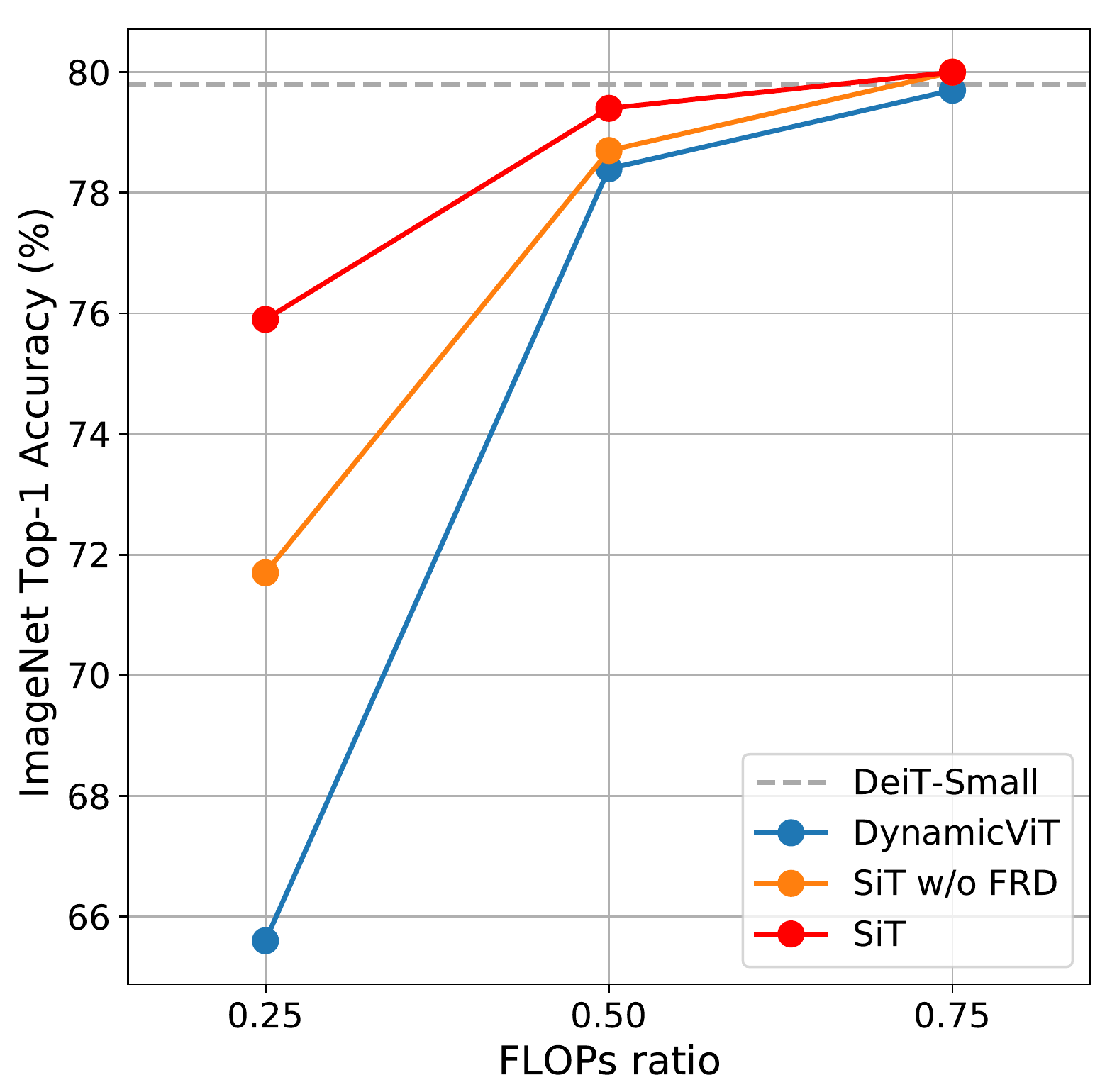}
        \subcaption{Our SiT achieves much higher accuracy than DynamicViT even without the distillation.}
        \label{comparision_dynamicvit}
    \end{minipage}
    \begin{minipage}[t]{0.44\linewidth}
        \centering
        \includegraphics[width=1\textwidth]{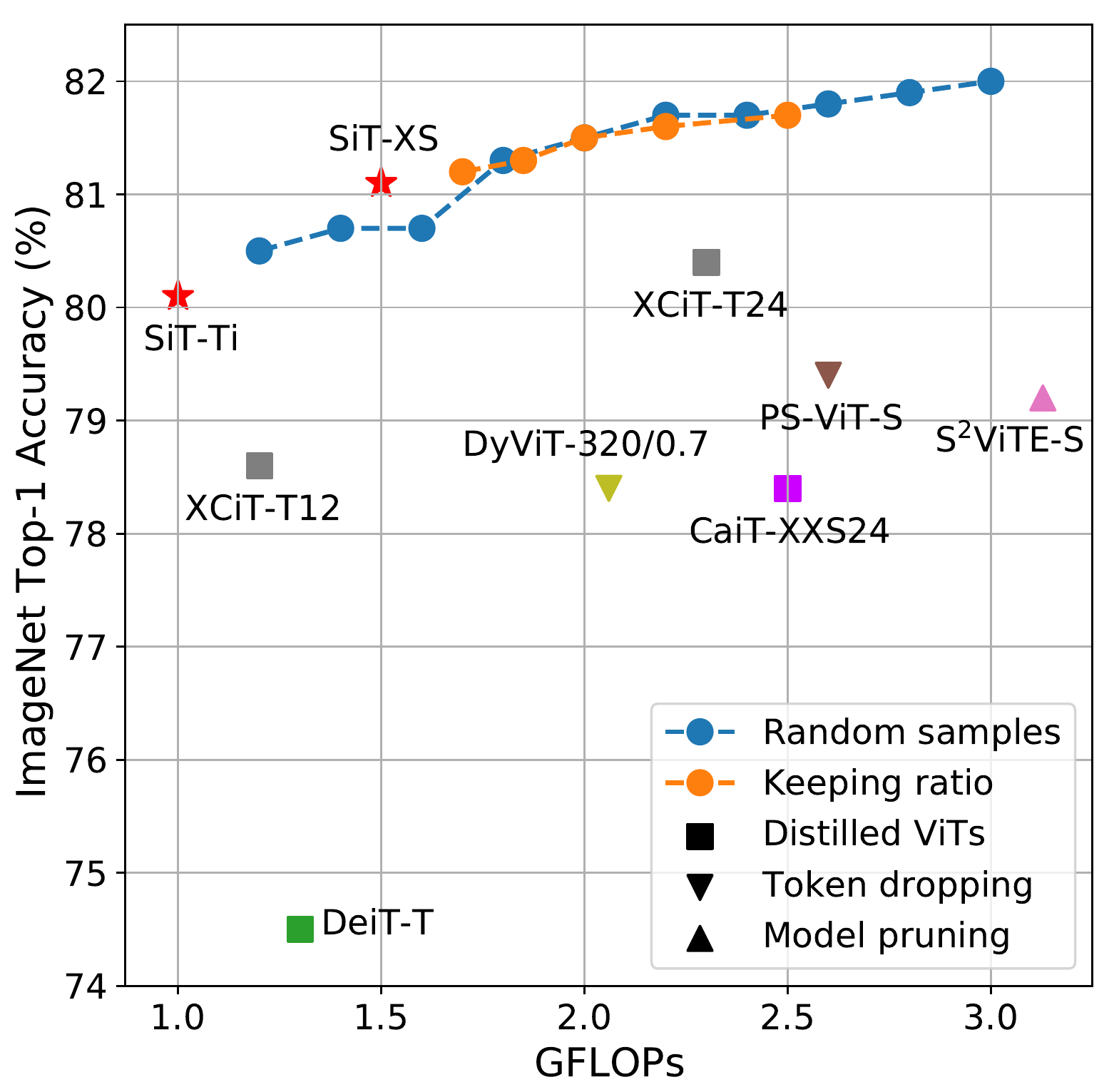}
        \subcaption{Our randomly sampled models consistently outperform other distilled and pruned ViTs.}
        \label{samples}
    \end{minipage}   
    \caption{\textbf{Effectiveness and robustness study.}
    We compare our SiT with the DynamicViT in Fig(a).
    To verify the robustness of our method, we randomly change the numbers of blocks in each stage (\ie, TSM location) and adjust the keeping ratio of TSM from 0.3 to 0.7 to sample a series of SiT-Ti models in Fig(b). 
    All the models are trained for 125 epochs without a CNN teacher.
    }
\end{figure*}

\subsection{Effectiveness and Robustness}

\noindent\textbf{Comparison to DynamicViT.}
In Fig. \ref{comparision_dynamicvit},
we compare our method with DynamicViT \cite{dynamicvit} on DeiT-S \cite{deit}.
When dropping too many tokens,
the performance of DynamicViT deteriorates dramatically.
Though it utilizes knowledge distillation to minimize the gap,
our SiT without the distillation consistently surpasses it under different FLOPs ratios,
especially under the smallest ratio.
Besides,
when armed with FRD,
our SiT can maintain performance better.

\noindent\textbf{TSM locations and keeping ratio.}
To verify the robustness of our method,
we conduct experiments on SiT-Ti as shown in Fig. \ref{samples}.
It clearly shows that all of the randomly sampled models outperform popular ViTs with knowledge distillation,
\eg, DeiT \cite{deit} and XCiT \cite{xcit}.
Besides,
compared with other counterparts based on token hard dropping \cite{dynamicvit,patchslimming} and structure pruning \cite{chasing},
our models surpass them by a large margin.
These results demonstrate our SiT is insensitive to the setting of TSM locations and keeping ratio.
To make a fair comparison with the state-of-the-art ViTs,
we set these hyper-parameters according to the FLOPs.

\input{compare_sota}

\begin{figure*}[tp]
    \centering
    \begin{minipage}[t]{0.44\linewidth}
        \centering
        \includegraphics[width=1\textwidth]{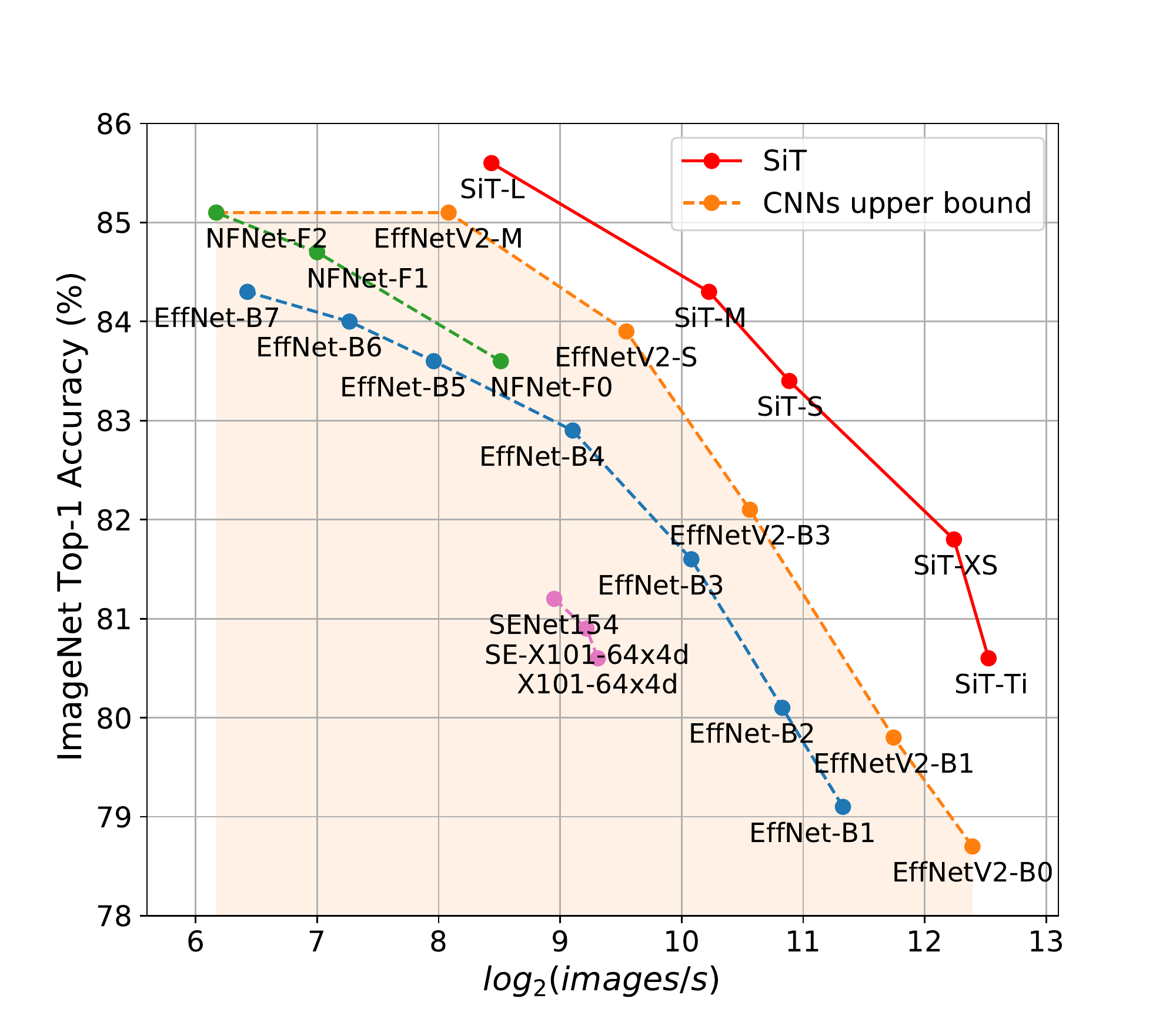}
        \subcaption{
        Comparison to CNNs.
        }
        \label{cnn_sota}
    \end{minipage}
    \begin{minipage}[t]{0.44\linewidth}
        \centering
        \includegraphics[width=1\textwidth]{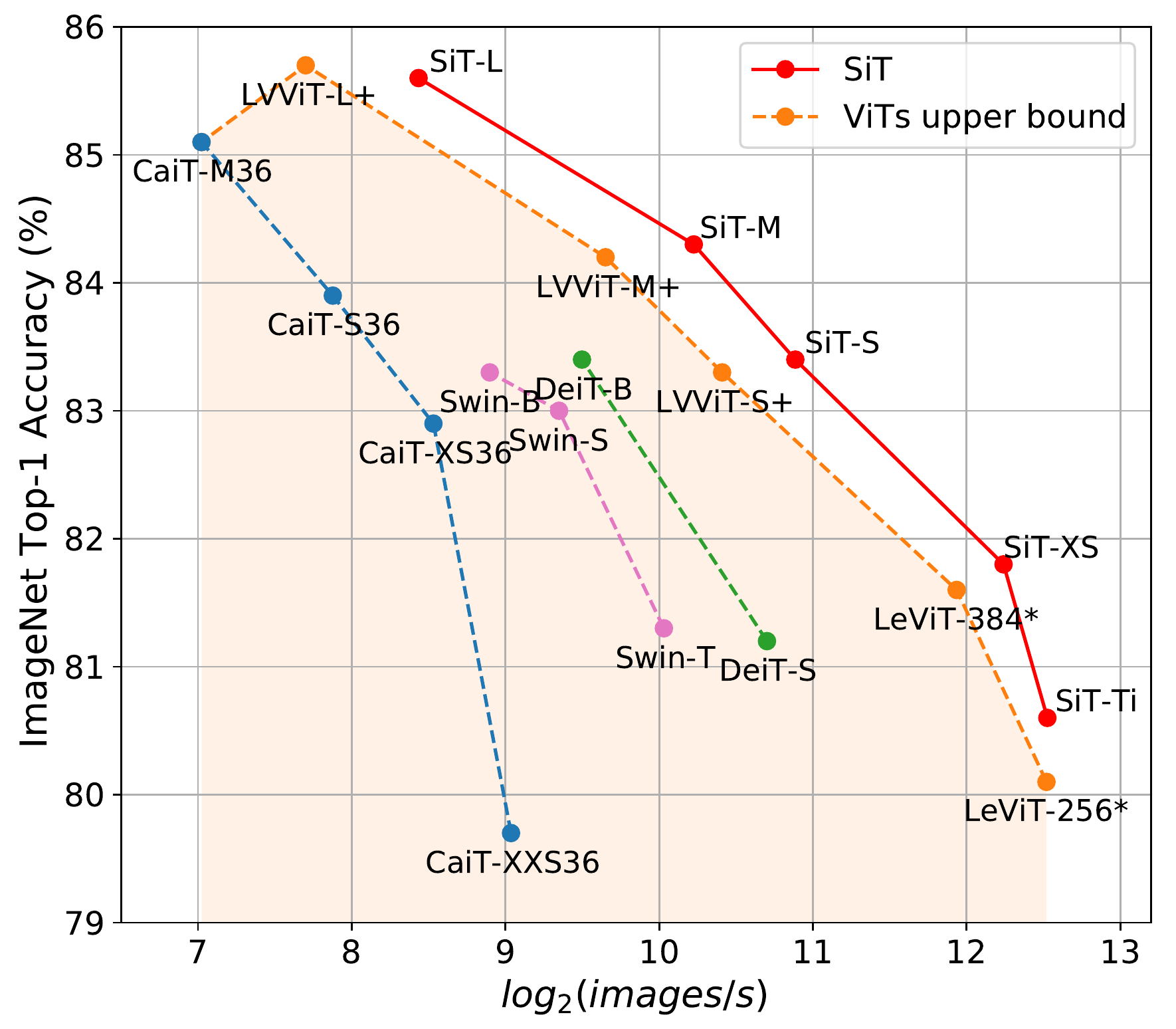}
        \subcaption{
        Comparison to ViTs.
        }
        \label{trans_sota}
    \end{minipage}
    \caption{\textbf{Speed \textit{vs.} Accuracy.}
    We compare our SiT with the previous state-of-the-art CNNs and ViTs in Fig(a) and Fig(b), respectively.
    ``LV-ViT+'' denotes our improved LV-ViT teacher.
    Our SiT surpasses the SOTA methods by a large margin, even the efficient models EfficientNetV2 \cite{efficientv2} and LeViT \cite{levit}.
    }
    \label{fig:curve_sota}
\end{figure*}

\subsection{Comparison to state-of-the-art}

In Table \ref{tab:sota_imagenet}, 
we compare SiT with other competitive CNNs and ViTs.
For a fair comparison,
we group these methods according to their top-1 accuracies.
The throughput is measured on a single 16GB V100 GPU under the same setting as LeViT \cite{levit}.
Our SiT-Ti is competitive with LeViT,
while the throughput is $\mathbf{3.2}\times$ than that of EfficientNet \cite{efficientnet}.
Note that EfficientNet is designed via extensive neural architecture search and LeViT is elaborately designed for fast inference.
For our larger model variants,
they perform better than EfficientNetV2 \cite{efficientv2} with simple training strategies.
Compared with the original LV-ViT \cite{lvvit},
our SiT is $\mathbf{1.5}\times$ faster than those with similar accuracy.

We further visualize the comparisons to the upper bounds of CNNs and ViTs in Fig. \ref{cnn_sota} and \ref{trans_sota}.
It clearly shows that our SiT achieves the best balance between throughput and accuracy,
surpassing the recent state-of-the-art CNNs and ViTs.

\subsection{Ablation Studies}
If not otherwise specified, 
all experiments for ablations are conducted on SiT-Ti and run with only 125 training epochs under the supervision of the original teacher model. 
``Token-MLP'' refers double linear layers along the token dimension.

\input{ablations}

\noindent
\textbf{Does token slimming outperform model scaling down?}
In Table \ref{ablation_efficiency},
we compare token slimming with model scaling down rules under the same computation limit.
For model scaling down,
we adapt the channel and depth individually.
Note that the above two models are trained from scratch for 300 epochs with token labeling \cite{lvvit}.
For token slimming,
we simply insert TSMs without FRD.
We also drop tokens and train it with extra distillation as in DynamicViT \cite{dynamicvit}.
It shows that scaling along the channel achieves higher accuracy than scaling along the depth but with lower throughput.
Besides,
token slimming can largely improve the throughput with higher performance.
However,
DynamicViT performs worse than our SiT without distillation,
since token hard dropping loses much discriminative information with a large slimming ratio.
Such results demonstrate simply inserting our TSM into vanilla ViT is able to achieve great performance. 

\noindent
\textbf{Does structure knowledge matter to self-slimmed learning?}
We further investigate whether the structure knowledge benefits the performance as shown in Table \ref{ablation_knowledge}.
For the teacher models,
we adopt different architectures (LV-ViT-S\cite{lvvit}, CaiT-S24\cite{cait}, and RegNetY-16GF\cite{regnet}) but similar accuracies for a fair comparison. 
It shows that training with the pre-trained weights for 125 epochs converges to higher results than those trained from scratch for 300 epochs.
Moreover,
we utilize structure knowledge via block-to-block mimicking,
which can further boost the performance.
It also reveals that higher similarity between students and teachers can bring greater improvements. 
\input{ablations2}
\input{ablations3}

\noindent
\textbf{Is self-slimmed learning robust to different FLOPs ratios?}
In Table \ref{ablation_ratio},
we empirically train models with different FLOPs ratios.
When the ratio is large than 0.5,
our FRD and CNN distillation are both helpful for maintaining performance.
However,
when the ratio is small,
CNN distillation leads to a higher performance drop,
while our FRD only drops the accuracy by 2.0\%.
These results demonstrate that our method is robust to different FLOPs ratios. 

\noindent
\textbf{Dynamic \textit{vs.} Static: Which aggregation manner works better for token slimming?}
To explore whether dynamic aggregation is better for token slimming,
we perform ablation experiments as shown in Table \ref{ablation_tsm}.
For static aggregation,
we choose different data-independent operations and maintain similar computation:
3$\times$3 average pooling/convolution with stride 2$\times$2,
and double linear layers with GELU function (``Token-MLP'').
It shows that learnable parameters are vital for token slimming since average pooling leads to a severe accuracy drop.
Besides,
the static aggregation methods with data-independent weights yield similar but inferior performance to our TSM (79.3\% \textit{vs.} 80.1\%).
Such comparisons prove that our TSM can generate more informative tokens.

\noindent
\textbf{Can contiguous upsampling recalibrate the features?}
We first recalibrate the original tokens by contiguous upsampling methods, \eg, bilinear interpolation and deconvolution.
As shown in Table \ref{ablation_rtsm},
these two spatial contiguous methods misalign the token relations and hurt the capacity compared with the baseline (without block-to-block mimicking).
In contrast,
``Token-MLP'' does not hurt the token representation,
and its accuracy can be further boosted to 80.1\% by the insertion of an MLP.

\noindent
\textbf{Does each distillation supervision help?}
Table \ref{ablation_mimic} presents that the soft logits supervision $\mathcal{L}_{\rm logits}$ brings 1.4\% accuracy gain.
When further introducing block-to-block knowledge supervision,
our model improves the accuracy by 1.1\%.
Finally, 
combining complementary hard label supervision,
the accuracy reaches 80.6\% with longer training epochs.

\noindent
\textbf{What are the appropriate loss weights?}
Table \ref{ablation_loss_weight} shows the settings of loss wights are robust in our SiT (trained for 100 epochs).
In fact,
we simply choose the weight of 2:2:1 to ensure different loss values are close in the early training.

\subsection{Visualization}

\begin{figure*}[tp]
    \centering
    \includegraphics[width=0.99\textwidth]{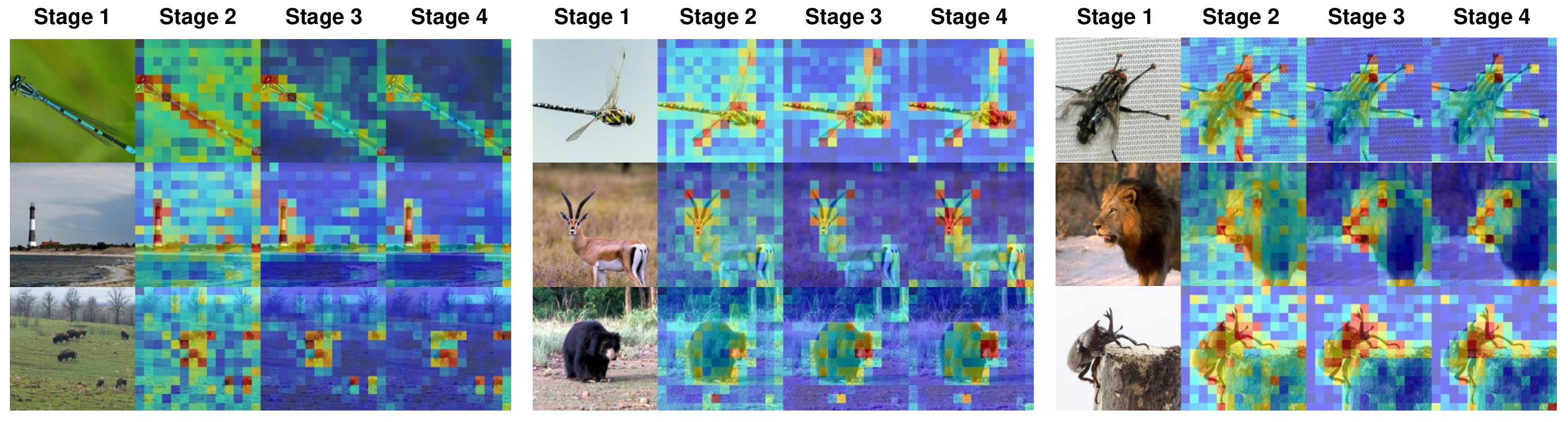}
    \caption{
   \textbf{Visualizations of our progressive token slimming.} 
   The blue/red tokens contribute less/more to the final informative tokens.
   Our method can zoom the attention scope to cover the key object,
   even with only 12.5\% of tokens.
   }
    \label{fig:vis_p1}
\end{figure*}

\noindent
\textbf{Qualitative token slimming visualization.}
Fig. \ref{fig:vis_p1} shows the original images and the token slimming procedure of our SiT-Ti.
We observe that the tokens of higher scores,
\ie,
brighter tokens,
are concentrated and tend to cover the key objects in the image.
It demonstrates our proposed TSM is able to localize the significant regions and predict accurate scores for the most informative tokens.


%% file: sota_imagenet.tex
\begin{table*}[t]
	\centering
    \caption{
    \textbf{Main results on ImageNet.}
    We apply our self-slimming learning on the state-of-the-art vanilla vision transformer LV-ViT \cite{lvvit}.
    $\ddagger$ means we adopt an extra CNN teacher.
    Our SiT can speed up LV-ViT $\mathbf{1.7}\times$ with a slight accuracy drop.
    For fast inference,
    our SiT can maintain 97\% of the performance while speeding up the original transformers by $\mathbf{3.6}\times$.
    }
    \begin{minipage}[t]{\linewidth}
        \centering
        \setlength\tabcolsep{3pt}
        \resizebox{0.85\linewidth}{!}{
    	\begin{tabular}{l|ccccc|cc}
    	    \Xhline{1.0pt}
    		\multirow{2}*{Model} & \multirow{2}*{Depth} & \multirow{2}*{Stage} &  \multirow{2}*{\makecell[c]{Embed \\ Dim}}  &  \multirow{2}*{Heads} & \multirow{2}*{Resolution} & \#Params & FLOPs \\
    		~ & ~ & ~ & ~ & ~ & ~ & (M) & (G) \\
    		\hline
    		SiT-Ti & 14 & \{1,1,1,11\} & 320 & 5 & 224$^2$ & 15.9 & 1.0 \\
    		SiT-XS & 16 & \{1,1,1,13\} & 384 & 6 & 224$^2$ & 25.6 & 1.5 \\
    		SiT-S & 16 & \{9,3,2,2\} & 384 & 6 & 224$^2$ & 25.6 & 4.0 \\
    		SiT-M & 20 & \{10,4,3,3\} & 512 & 8 & 224$^2$ & 55.6 & 8.1 \\
    		SiT-L & 24 & \{10,4,3,7\} & 768 & 12 & 288$^2$ & 148.2 & 34.4 \\
            \Xhline{1.0pt}
    	\end{tabular}
        }
        \subcaption{Model architecture settings}
        \label{abaltion_efficiency}
    \end{minipage}
    \vspace{-0.6cm}
    \begin{minipage}[t]{\linewidth}
        \centering
        \setlength\tabcolsep{4pt}
        \resizebox{0.85\textwidth}{!}{
    	\begin{tabular}{l|ccc|cc}
    	    \Xhline{1.0pt}
    		\multirow{3}*{Model} & \multicolumn{3}{c|}{Student} & \multicolumn{2}{c}{Teacher} \\
    		~ & Throughput & Top-1 & Top-1$\ddagger$ & Throughput & Top-1 \\
    		~ & (image/s) & (\%) & (\%) & (image/s) & (\%) \\
    		\hline
    		SiT-Ti & 5896 \color{blue}{($\mathbf{3.2}\times$)} & 80.1 ($-$2.0) & 80.6 ($-$1.5) & 1827 & 82.1 \\
    		SiT-XS & 4839 \color{blue}{($\mathbf{3.6}\times$)} & 81.1 ($-$2.2) & 81.8 ($-$1.5) & 1360 & 83.3 \\
    		SiT-S & 1892 ($\mathbf{1.4}\times$) & 83.2 \color[RGB]{17, 122, 101}{(\textbf{$-$0.1})} & 83.4 ($+$0.1) & 1360 & 83.3 \\
    		SiT-M & 1197 ($\mathbf{1.5}\times$) & 84.1 \color[RGB]{17, 122, 101}{(\textbf{$-$0.1})} & 84.3 ($+$0.1) & 804 & 84.2 \\
    		SiT-L & 346 ($\mathbf{1.7}\times$) & 85.6 \color[RGB]{17, 122, 101}{(\textbf{$-$0.1})} & - & 204 & 85.7 \\
            \Xhline{1.0pt}
    	\end{tabular}
        }
        \subcaption{Efficiency comparisons}
        \label{abaltion_efficiency}
    \end{minipage}
    \vspace{-0.4cm}
    \label{main_results}
\end{table*}

%% file: compare_sota.tex
\begin{table*}[t]
	\centering
    \caption{
    \textbf{Comparison to the state-of-the-art on ImageNet.} 
    The models marked in \textit{gray} color are trained with distillation supervision from a powerful CNN for 300 epochs.
    Our SiT achieves the best performance trade-off.
    }
    \setlength\tabcolsep{1pt}
    \resizebox{0.77\textwidth}{!}{
    	\begin{tabular}{l|cccc|c}
            \Xhline{1.0pt}
    		\multirow{2}*{Model} & \multirow{2}*{Resolution} & \#Params & FLOPs & Throughput & {ImageNet} \\
    		~ & ~ & (M) & (G) & (image/s) & Top-1(\%) \\
    		\hline
     		EfficientNet-B1 \cite{efficientnet} & $240^2$ & 7.8 & 0.7 & 2559 & 79.1 \\
    		EfficientNet-B2 \cite{efficientnet} & $260^2$ & 9.1 & 1.1 & 1808 & 80.1 \\
     		\cellcolor{gray!20}DeiT-T \cite{deit} & \cellcolor{gray!20}$224^2$ & \cellcolor{gray!20}5.9 & \cellcolor{gray!20}1.3 & \cellcolor{gray!20}3346 & \cellcolor{gray!20}74.5 \\
    		\cellcolor{gray!20}LeViT-256 \cite{levit} & \cellcolor{gray!20}$224^2$ & \cellcolor{gray!20}18.9 & \cellcolor{gray!20}1.1 & \cellcolor{gray!20}5802 & \cellcolor{gray!20}80.1 \\
    		\textbf{SiT-Ti} & $224^2$ & 15.9 & 1.0 & \textbf{5896} & 80.1 \\
    		\cellcolor{gray!20}\textbf{SiT-Ti} & \cellcolor{gray!20}$224^2$ & \cellcolor{gray!20}16.2 & \cellcolor{gray!20}1.0 & \cellcolor{gray!20}5833 & \cellcolor{gray!20}\textbf{80.6} \\
    		\Xhline{0.7pt}
    		EfficientNet-B3 \cite{efficientnet} & $300^2$ & 12.2 & 1.9 & 1062 & 81.6 \\
    		Swin-T \cite{swin} & $224^2$ & 28.3 & 4.5 & 1023 & 81.3 \\
    		\cellcolor{gray!20}DeiT-S \cite{deit} & \cellcolor{gray!20}$224^2$ & \cellcolor{gray!20}22.4 & \cellcolor{gray!20}4.6 & \cellcolor{gray!20}1598 & \cellcolor{gray!20}81.2 \\
    		\cellcolor{gray!20}LeViT-384 \cite{levit} & \cellcolor{gray!20}$224^2$ & \cellcolor{gray!20}39.1 & \cellcolor{gray!20}2.4 & \cellcolor{gray!20}3876 & \cellcolor{gray!20}81.6 \\
    		\textbf{SiT-XS} & $224^2$ & 25.6 & 1.5 & \textbf{4839} & 81.1 \\
    		\cellcolor{gray!20}\textbf{SiT-XS} & \cellcolor{gray!20}$224^2$ & \cellcolor{gray!20}26.0 & \cellcolor{gray!20}1.5 & \cellcolor{gray!20}4798 & \cellcolor{gray!20}\textbf{81.8} \\
    		\Xhline{0.7pt}
    		EfficientNet-B4 \cite{efficientnet} & $380^2$ & 19.3 & 4.6 & 545 & 82.9 \\
    		Swin-B \cite{swin} & $224^2$ & 87.8 & 15.5 & 474 & 83.3 \\
    		\cellcolor{gray!20}DeiT-B \cite{deit} & \cellcolor{gray!20}$224^2$ & \cellcolor{gray!20}87.3 & \cellcolor{gray!20}17.7 & \cellcolor{gray!20}718 & \cellcolor{gray!20}83.4 \\
    		\cellcolor{gray!20}LV-ViT-S \cite{lvvit} & \cellcolor{gray!20}$224^2$ & \cellcolor{gray!20}26.2 & \cellcolor{gray!20}6.6 & \cellcolor{gray!20}1270 & \cellcolor{gray!20}83.3 \\
    		\textbf{SiT-S} & $224^2$ & 25.6 & 4.0 & \textbf{1892} & 83.2 \\
    		\cellcolor{gray!20}\textbf{SiT-S} & \cellcolor{gray!20}$224^2$ & \cellcolor{gray!20}26.0 & \cellcolor{gray!20}4.0 & \cellcolor{gray!20}1873 & \cellcolor{gray!20}\textbf{83.4} \\
    		\Xhline{0.7pt}
    		EfficientNet-B6 \cite{efficientnet} & $528^2$ & 43.0 & 19.9 & 153 & 84.0 \\
    		EfficientNetV2-S \cite{efficientv2} & $384^2$ & 21.5 & 8.5 & 742 & 83.9 \\
     		\cellcolor{gray!20}CaiT-S36 \cite{cait} & \cellcolor{gray!20}$224^2$ & \cellcolor{gray!20}68.2 & \cellcolor{gray!20}13.9 & \cellcolor{gray!20}233 & \cellcolor{gray!20}83.9 \\
    		\cellcolor{gray!20}LV-ViT-M \cite{lvvit} & \cellcolor{gray!20}$224^2$ & \cellcolor{gray!20}55.8 & \cellcolor{gray!20}12.7 & \cellcolor{gray!20}768 & \cellcolor{gray!20}84.1 \\
    		\textbf{SiT-M} & $224^2$ & 55.6 & 8.1 & \textbf{1197} & 84.1 \\
    		\cellcolor{gray!20}\textbf{SiT-M} & \cellcolor{gray!20}$224^2$ & \cellcolor{gray!20}56.2 & \cellcolor{gray!20}8.1 & \cellcolor{gray!20}1185 & \cellcolor{gray!20}\textbf{84.3} \\
    		\Xhline{0.7pt}
    		EfficientNetV2-M \cite{efficientv2} & $480^2$ & 54.1 & 25.0 & 271 & 85.1 \\
     		NFNet-F1 \cite{nfnet} & $320^2$ & 132.6 & 36.0 & 128 & 84.7 \\
    		\cellcolor{gray!20}CaiT-M36 \cite{cait} & \cellcolor{gray!20}$224^2$ & \cellcolor{gray!20}270.1 & \cellcolor{gray!20}53.4 & \cellcolor{gray!20}130 & \cellcolor{gray!20}85.1 \\
    		\cellcolor{gray!20}LV-ViT-L \cite{lvvit} & \cellcolor{gray!20}$288^2$ & \cellcolor{gray!20}150.1 & \cellcolor{gray!20}58.8 & \cellcolor{gray!20}208 & \cellcolor{gray!20}85.3 \\
    		\textbf{SiT-L} & $288^2$ & 148.2 & 34.4 & \textbf{346} & \textbf{85.6} \\
            \Xhline{1.0pt}
    	\end{tabular}
    }
    \label{tab:sota_imagenet}
    \vspace{-0.3cm}
\end{table*}

%% file: ablations.tex
\begin{table*}[t]
    \begin{minipage}[t]{0.46\linewidth}
        \centering
        \caption{Efficiency comparison. 
        }
        \setlength\tabcolsep{2.5pt}
        \resizebox{\textwidth}{!}{
        \begin{tabular}[t]{l|c|c}
            \Xhline{1.0pt}
            Method & Top-1 & Throughput \\
            \hline
            Structure-width & 76.3 & 2947 \\
            \hline
            Structure-depth & 69.4 & 5652 \\
            \hline
            DynamicViT\cite{dynamicvit} & 75.7 & 5762 \\
            \hline
            SiT w/o FRD & \textbf{77.7} & \textbf{5896} \\
            \Xhline{1.0pt}
        \end{tabular}
        }
        \label{ablation_efficiency}
    \end{minipage}
    \hspace{1mm}
    \begin{minipage}[t]{0.52\linewidth}
        \centering
        \caption{Inherited knowledge.
        }
        \resizebox{\textwidth}{!}{
        \begin{tabular}[t]{l|c|c|c}
            \Xhline{1.0pt}
            \multirow{2}*{Knowledge} & Self & CaiT & RegNet \\
            ~ & 83.3 & 83.5 & 82.9 \\
            \hline
            Scratch & \textbf{80.1} & 79.9 & 79.2 \\
            \hline
            Fine-tuning & \textbf{80.5} & 80.2 & 80.0 \\
            \hline
            Fine-tuning+Structure  & \textbf{81.1} & 80.6 & 80.2 \\
            \Xhline{1.0pt}
        \end{tabular}
        }
        \label{ablation_knowledge}
    \end{minipage}
    \vspace{-0.6cm}
\end{table*}

%% file: ablations2.tex
\begin{table*}[t]
    \begin{minipage}[t]{0.3\linewidth}
        \centering
        \caption{Robustness of slimming ratios.
        }
        \resizebox{\textwidth}{!}{
        \begin{tabular}[t]{l|c|c}
            \Xhline{1.0pt}
            Ratio& $\mathcal{L}_{\rm logits}$+$\mathcal{L}_{\rm token}$ & $\mathcal{L}_{\rm hard}$ \\
            \hline
            1 & 82.1 & 82.1 \\
            \hline
            0.75 & 82.0 & 82.0 \\
            \hline
            0.5 & 81.6 & 81.3 \\
            \hline
            0.25 & 80.1 & 78.4 \\
            \Xhline{1.0pt}
        \end{tabular}
        }
        \label{ablation_ratio}
    \end{minipage}
    \hspace{0.5mm}
    \begin{minipage}[t]{0.345\linewidth}
        \centering
        \caption{Slimming.
        }
        \resizebox{\textwidth}{!}{
        \begin{tabular}[t]{l|c|c}
            \Xhline{1.0pt}
            Method & GFLOPs & Top-1 \\
            \hline
            Baseline & 3.5 & 82.1 \\
            \hline
            3$\times$3 AvgPool & 1.0 & 77.4 \\
            \hline
            3$\times$3 Conv & 1.0 & 79.3 \\
            \hline
            Token-Mixer & 1.1 & 79.3 \\
            \hline
            Our TSM & \textbf{1.0} & \textbf{80.1} \\
            \Xhline{1.0pt}
        \end{tabular}
        }
        \label{ablation_tsm}
    \end{minipage}
    \hspace{0.5mm}
    \begin{minipage}[t]{0.307\linewidth}
        \centering
        \caption{Recalibration.
        }
        \setlength\tabcolsep{7pt}
        \resizebox{\textwidth}{!}{
        \begin{tabular}[t]{l|c}
            \Xhline{1.0pt}
            Method & Top-1 \\
            \hline
            Baseline & 79.0 \\
            \hline
            Interpolation & 78.3 \\
            \hline
            Deconvolution & 78.4 \\
            \hline
            Token-MLP & 79.0 \\
            \hline
            Our RTSM & \textbf{80.1} \\
            \Xhline{1.0pt}
        \end{tabular}
        }
        \label{ablation_rtsm}
    \end{minipage}
    \vspace{-0.7cm}
\end{table*}

%% file: ablations3.tex
\begin{table*}[t]
    \begin{minipage}[t]{0.6\linewidth}
        \centering
        \caption{Knowledge distillation. 
        }
        \setlength\tabcolsep{4pt}
        \resizebox{1\textwidth}{!}{
        \begin{tabular}[t]{l|c}
            \Xhline{1.0pt}
            Method & Top-1 \\
            \hline
            Baseline & 77.7 \\
            \hline
            +$\mathcal{L}_{\rm logits}$ & 79.0\color[RGB]{17, 122, 101}{$\mathbf{(+1.3)}$}\\
            \hline
            +$\mathcal{L}_{\rm logits}$+$\mathcal{L}_{\rm token}$ & 80.1\color[RGB]{17, 122, 101}{$\mathbf{(+2.4)}$} \\
            \hline
            +$\mathcal{L}_{\rm logits}$+$\mathcal{L}_{\rm token}$+$\mathcal{L}_{\rm hard}$ & 80.2\color[RGB]{17, 122, 101}{$\mathbf{(+2.5)}$} \\
            \hline
            +$\mathcal{L}_{\rm logits}$+$\mathcal{L}_{\rm token}$+$\mathcal{L}_{\rm hard}$+Longer training & 80.6\color[RGB]{17, 122, 101}{$\mathbf{(+2.9)}$} \\
            \Xhline{1.0pt}
        \end{tabular}
        }
        \label{ablation_mimic}
    \end{minipage}
    \hspace{1mm}
    \begin{minipage}[t]{0.37\linewidth}
        \centering
        \caption{Loss weights.}
        \setlength\tabcolsep{9.5pt}
        \resizebox{\textwidth}{!}{
        \begin{tabular}[t]{c|c}
            \Xhline{1.0pt}
            $\lambda_{token}$:$\lambda_{logit}$:$\lambda_{hard}$ & Top-1 \\
            \hline
            1:1:1 & 79.3 \\
            \hline
            1:2:1 & 79.4 \\
            \hline
            1:2:2 & 79.5 \\
            \hline
            2:1:1 & \textbf{79.6} \\
            \hline
            2:2:1 & \textbf{79.6} \\
            \Xhline{1.0pt}
        \end{tabular}
        }
        \label{ablation_loss_weight}
    \end{minipage}
    \vspace{-0.6cm}
\end{table*}

%% file: 5_conclusions.tex
\section{Conclusions}
In this paper,
we propose a generic self-slimmed learning method for vanilla vision transformers (SiT),
which can speed up the ViTs with negligible accuracy drop.
Our concise TSM softly integrates redundant tokens into fewer informative ones.
For stable and efficient training,
we introduce a novel FRD framework to leverage structure knowledge,
which can densely transfer token information in a flexible auto-encoder manner.
Extensive experiments demonstrate the effectiveness of our SiT.
By simply arming LV-ViT with our SiT, 
we achieve new state-of-the-art performance on ImageNet, 
surpassing recent CNNs and ViTs.

\noindent
\textbf{Acknowledgements.}
This work is partially supported by National Key R\&D Program of China under Grant 2019YFB2102400, National Natural Science Foundation of China (61876176), the Joint Lab of CAS-HK, Shenzhen Institute of Artificial Intelligence and Robotics for Society, the Shanghai Committee of Science and Technology (Grant No. 21DZ1100100).


%% file: X_supplementary.tex
\appendix



\section{More details about teacher models}
\input{teacher_model}
Table \ref{teacher_models} shows more details about our teacher models.
We elaborately design different patch stems for our LV-ViT \cite{lvvit} models.

\section{More robustness analysis.}
\input{more_robustness}
We conduct more analysis based on our LV-ViT-S in Table \ref{more_robustness}.
It shows that our self-slimmed learning is also robust to different FLOPs ratios on LV-ViT-S.
Moreover,
our method still performs better than CNN distillation on larger model.

\clearpage
\section{More experiments on DeiT}
\input{results_deit}
We also verify the effectiveness of our self-slimmed learning on DeiT as illustrated in Table \ref{tab:deit}. 
For the FLOPs ratio of 0.5 and 0.25, 
the stage numbers are \{3,4,3,2\} and \{1,1,1,9\} respectively.
Specifically, 
we conduct the experiments on the original DeiT \cite{deit} and its variant with lightweight convolutional patch embedding. 
Both models achieve similar accuracy with the same computational costs.
However, 
we observe the performance of their students is quite different especially at a small FLOPs ratio.
DeiT$_{P}$ suffers severe performance deterioration when $75\%$ computation is reduced,
while DeiT$_{C}$ only drops the accuracy by 2.5\%.
More importantly, 
DeiT$_{C}$ generally obtain higher accuracies than DeiT$_{P}$ at a relatively higher FLOPs ratio.
It demonstrates that the models with convolutional patch embedding are more redundant and friendly to slimming.
In addition, 
we also compare our DKD with the CNN distillation under different settings.
The layer-to-layer dense knowledge distillation consistently brings more performance gains than CNN distillation.
It is worth mentioning that,
self-slimmed learning is also complementary to the extra CNN distillation.
Surprisingly, the best student model of DeiT$_{C}$ even outperforms the teacher by $0.6\%$ top-1 accuracy while running $2\times$ faster under the joint supervision.
These results prove the effectiveness and generalization ability of our self-slimmed learning.

\input{compare_dvt}

As described in Table \ref{tab:dynamic}, we further compare our self-slimmed learning with the recent method,
\ie,
DynamicViT.
We observe that our SiT runs slightly faster than DynamicViT with the same FLOPs,
which reveals our TSM presents better inference efficiency than the prediction module of DynamicViT.
More importantly, 
thanks to the soft-slimming designs, 
SiT outperforms DynamicViT by a large margin (5.3\%-10.0\%) at the FLOPs ratio of 0.25.
For the large FLOPs ratio,
our SiT still obtains at least 0.7\% higher accuracy than DynamicViT, 
proving the soft slimming triumphs the hard dropping manner.

\section{More experiments on Swin Transformer}
\input{swin}
\noindent
Note that the recent slimming methods \cite{dynamicvit} only works for vanilla ViTs.
Since the hierarchical ViTs generally introduce structured operations like convolution and relative position bias,
it's not suitable for arbitrary token dropping.
To verify the generality of our SiT,
we adapt the typical hierarchical network (i.e., Swin) with SiT,
modifying some of the structured operations.
Table \ref{tab:swin} shows that arming Swin with SiT,
we can also improve its throughput without accuracy drop.
We will focus on more elegant token slimming method in the future.

\begin{figure*}[tp]
    \centering
    \includegraphics[width=0.99\textwidth]{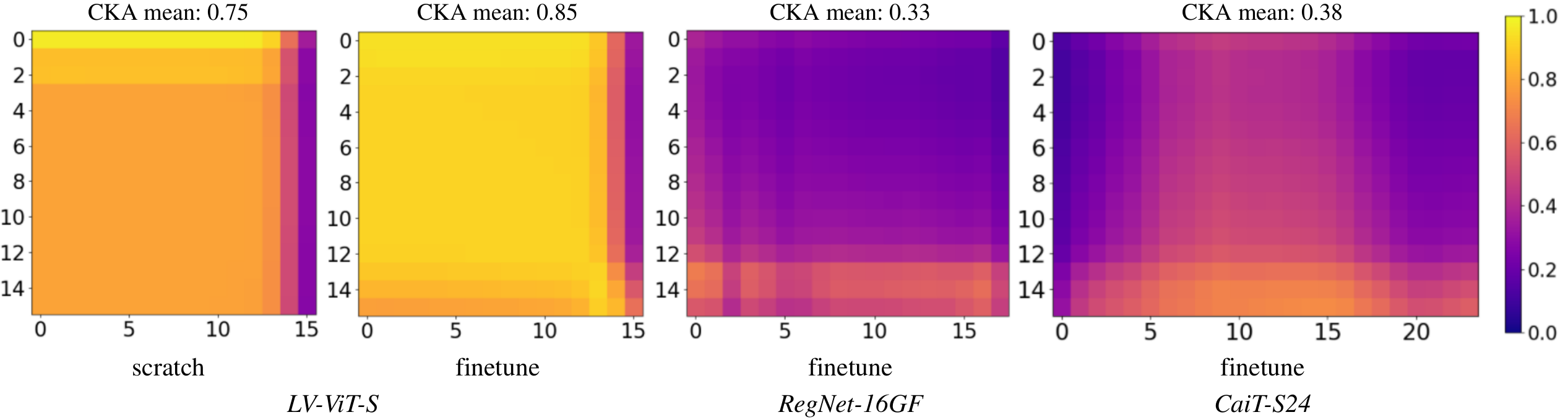}
    \caption{
    \textbf{Cross CKA heatmap between different student models and the teacher models.}
    We adopt LV-ViT-S \cite{lvvit} as student. Transfering knowledge densely from same structure yields the largest similarity.
    }
    \label{fig:cka}
\end{figure*}

\section{More visualizations}
\begin{figure*}[tp]
    \centering
    \includegraphics[width=1.0\textwidth]{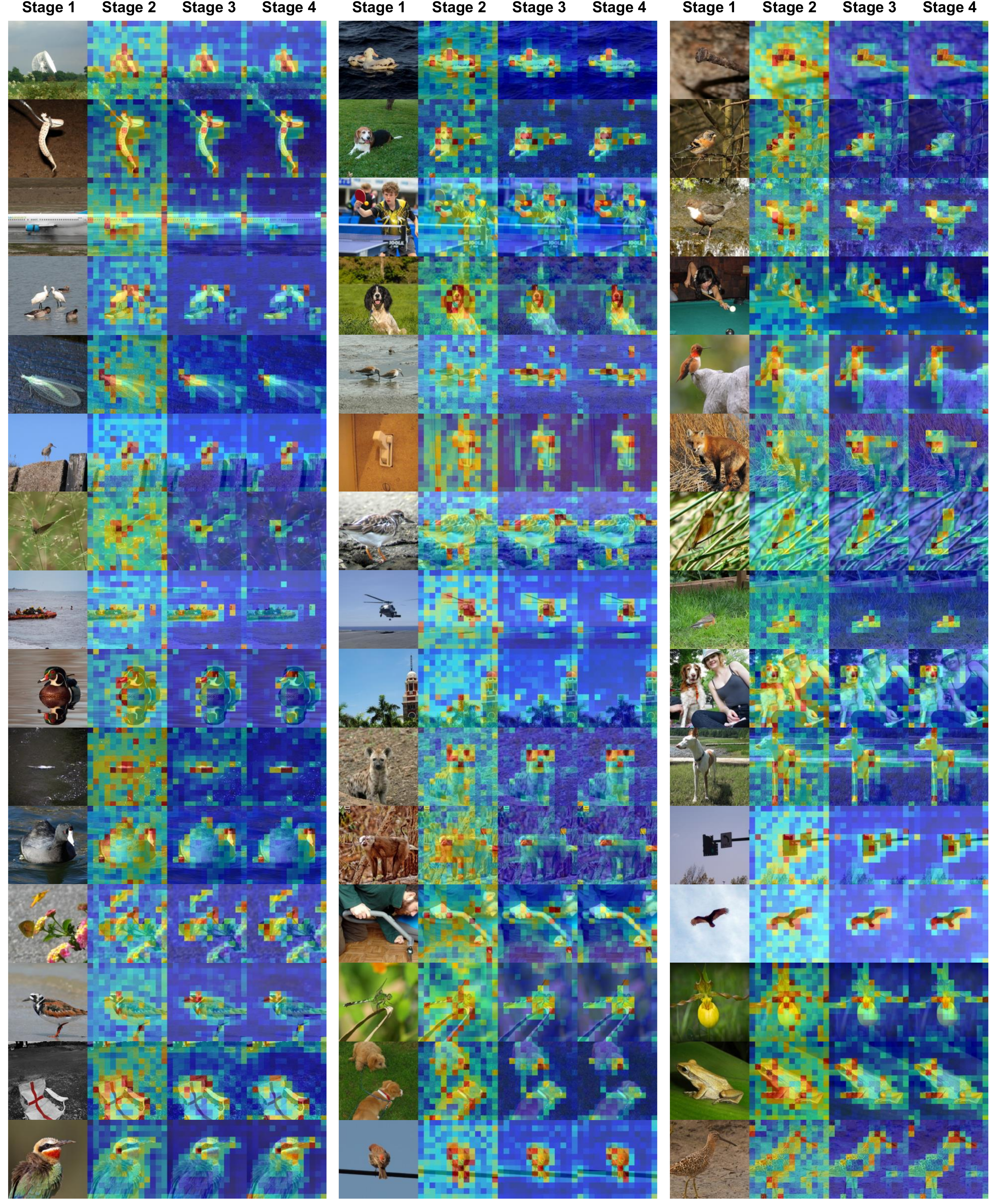}
    \caption{
    More visualizations of our SiT.}
    \label{fig:vis}
\end{figure*}

\noindent
\textbf{Qualitative token slimming visualization.}
We present more visualizations of our progressive token slimming in Figure \ref{fig:vis}.

\noindent
\textbf{Qualitative FRD visualization.}
In Fig. \ref{fig:cka},
we compute the CKA \cite{cka} heatmap by comparing all layers of the student models (LV-ViT-S) with all layers of their teacher models.
It shows that the CKA similarities between the similar structures are generally higher than those between different structures (0.75/0.85 \textit{vs.} 0.33/0.38).
Interestingly, we find the pre-trained weights inherited by the student force itself to be similar to its teacher.
Besides,
for similar structures,
the CKA similarities in the shallow layers are higher than those in deep layers.
It is mainly because we slim a large number of tokens after the third layer,
leading to an inevitable information loss.
As for different structures,
the CKA similarities in the deep layers are higher than those in shallow layers,
which is mainly because the logits distillation provides direct supervision for features in the deeper layers.
Note that the above observations are consistent with the results in our experiments,
which reveals that teachers with similar structures can transfer structure knowledge better for higher performance.

%% file: teacher_model.tex
\begin{table}[tp]
	\centering
    \caption{Details of teacher models. 
    The head dimensions of all the models are set to 64.
    `ks' means kernel size.
    `st' means stride.
    `oc' means output channel number.}
    \setlength\tabcolsep{2pt}
    \resizebox{0.8\textwidth}{!}{
    	\begin{tabular}{c|c|c|c|c|c}
            \Xhline{0.8pt}
    		\multirow{2}*{Teacher} & \multirow{2}*{Student} & \multirow{2}*{Resolution} & \multirow{2}*{Depth} &  \multirow{2}*{Heads} & 
    		Patch Stem\\
    		~ & ~ & ~ & ~ & ~ & (ks, st, oc) \\
    		\hline
    		~ & \multirow{4}*{SiT-Ti} & \multirow{4}*{224$^2$} & \multirow{4}*{14} & \multirow{4}*{5} & (3$\times$3, 2$\times$2, 40)\\
    		Our & ~ & ~ & ~ & ~ & (3$\times$3, 2$\times$2, 80)\\
    	    LV-ViT-Ti & ~ & ~ & ~ & ~ & (3$\times$3, 2$\times$2, 160)\\
    		~ & ~ & ~ & ~ & ~ & (3$\times$3, 2$\times$2, 320)\\
    		\hline
    		~ & ~ & \multirow{4}*{224$^2$} & \multirow{4}*{16} & \multirow{4}*{6} & (3$\times$3, 2$\times$2, 48)\\
    		Our & SiT-XS & ~ & ~ & ~ & (3$\times$3, 2$\times$2, 96)\\
    	    LV-ViT-S & SiT-S & ~ & ~ & ~ & (3$\times$3, 2$\times$2, 192)\\
    		~ & ~ & ~ & ~ & ~ & (3$\times$3, 2$\times$2, 384)\\
    		\hline
    		~ & \multirow{4}*{SiT-M} & \multirow{4}*{224$^2$} & \multirow{4}*{20} & \multirow{4}*{8} & (3$\times$3, 2$\times$2, 64)\\
    		Our & ~ & ~ & ~ & ~ & (3$\times$3, 2$\times$2, 128)\\
    	    LV-ViT-M & ~ & ~ & ~ & ~ & (3$\times$3, 2$\times$2, 256)\\
    		~ & ~ & ~ & ~ & ~ & (3$\times$3, 2$\times$2, 512)\\
    		\hline
    		~ & \multirow{6}*{SiT-L} & \multirow{6}*{288$^2$} & \multirow{6}*{24} & \multirow{6}*{12} & (3$\times$3, 2$\times$2, 96)\\
    		~ & ~ & ~ & ~ & ~ & (3$\times$3, 1$\times$1, 96)\\
    		Our & ~ & ~ & ~ & ~ & (3$\times$3, 1$\times$1, 96)\\
    		LV-ViT-L & ~ & ~ & ~ & ~ & (3$\times$3, 2$\times$2, 192)\\
    	    ~ & ~ & ~ & ~ & ~ & (3$\times$3, 2$\times$2, 384)\\
    		~ & ~ & ~ & ~ & ~ & (3$\times$3, 2$\times$2, 768)\\
    		\hline
            \Xhline{0.8pt}
    	\end{tabular}
    }
    \label{teacher_models}
    \vspace{-0.3cm}
\end{table}

%% file: more_robustness.tex
\begin{table}[tp]
	\centering
    \caption{Robustness analysis based on our LV-ViT-S.}
    \setlength\tabcolsep{16pt}
    \resizebox{0.6\textwidth}{!}{
    	\begin{tabular}[t]{l|c|c}
            \Xhline{1.0pt}
            Ratio& $\mathcal{L}_{\rm logits}$+$\mathcal{L}_{\rm token}$ & $\mathcal{L}_{\rm hard}$ \\
            \hline
            1 & 83.3 & 83.3 \\
            \hline
            0.75 & 83.2 & 83.0 \\
            \hline
            0.5 & 82.6 & 82.2 \\
            \hline
            0.25 & 80.9 & 80.0 \\
            \Xhline{1.0pt}
        \end{tabular}
    }
    \label{more_robustness}
    \vspace{-0.3cm}
\end{table}

%% file: results_deit.tex
\begin{table}[tp]
	\centering
    \caption{More results on DeiT. ``DeiT$_{P}$'' indicates the original DeiT and ``DeiT$_{C}$'' refers to the variant with lightweight convolutional patch embedding stacked by four 3$\times$3 convolutions (2$\times$2 stride) and one point-wise convolution.
    }
    \setlength\tabcolsep{2pt}
    \resizebox{0.8\textwidth}{!}{
    	\begin{tabular}{l|cc|cc|cc}
    		\Xhline{1.0pt}
    		\multirow{2}*{Model} & FLOPs & FLOPs &  $\mathcal{L}_{\rm logits}$ & \multirow{2}*{$\mathcal{L}_{\rm hard}$} & Throughput & {ImageNet} \\
    		~ & ratio & (G) & +$\mathcal{L}_{\rm token}$ & ~ & (image/s) & Top-1(\%) \\
    		\hline
    		\multirow{13}*{DeiT$_{P}$-S} 
    		& \multirow{4}*{0.25} 
    		& 1.1 & \color{lightgray}{\XSolidBrush}  &  \color{lightgray}{\XSolidBrush} & 6413\color{blue}{($\mathbf{3.9}\times$)} & 71.6(-8.2) \\
    		~ & ~ & 1.1 & \CheckmarkBold &  \color{lightgray}{\XSolidBrush} & 6413\color{blue}{($\mathbf{3.9}\times$)}  & 75.9(-3.9) \\
    		~ & ~ & 1.1 & \color{lightgray}{\XSolidBrush} & \CheckmarkBold & 6286\color{blue}{($\mathbf{3.8}\times$)}  & 72.9(-6.9) \\
    		~ & ~ & 1.1 & \CheckmarkBold & \CheckmarkBold & 6286\color{blue}{($\mathbf{3.8}\times$)} & 75.3(-4.5) \\
    		\cline{2-7}
    		~ & \multirow{4}*{0.5} 
    		& 2.3 & \color{lightgray}{\XSolidBrush}  &  \color{lightgray}{\XSolidBrush} & 3308($\mathbf{2.0}\times$) & 78.6\color[RGB]{17, 122, 101}{$\mathbf{(-1.3)}$} \\
    		~ & ~ & 2.3 & \CheckmarkBold & \color{lightgray}{\XSolidBrush} & 3308($\mathbf{2.0}\times$) & 79.4\color[RGB]{17, 122, 101}{$\mathbf{(-0.4)}$} \\
    		~ & ~ & 2.3 & \color{lightgray}{\XSolidBrush} & \CheckmarkBold & 3262($\mathbf{2.0}\times$) & 78.8\color[RGB]{17, 122, 101}{$\mathbf{(-1.0)}$} \\
    		~ & ~ & 2.3 & \CheckmarkBold & \CheckmarkBold & 3262($\mathbf{2.0}\times$) & 79.8\color[RGB]{17, 122, 101}{$\mathbf{(+0.0)}$} \\
    		\cline{2-7}
    		~ & 1
    		& 4.6 & \color{lightgray}{\XSolidBrush}  &  \color{lightgray}{\XSolidBrush} & 1637 & 79.8 \\
    		\hline

    		\multirow{13}*{DeiT$_{C}$-S}
    		& \multirow{4}*{0.25} 
    		& 1.1 & \color{lightgray}{\XSolidBrush}  &  \color{lightgray}{\XSolidBrush} & 5898\color{blue}{($\mathbf{3.7}\times$)} & 76.1(-3.9) \\
    		~ & ~ & 1.1 & \CheckmarkBold &  \color{lightgray}{\XSolidBrush} & 5898\color{blue}{($\mathbf{3.7}\times$)}  & 78.4(-1.6) \\
    		~ & ~ & 1.1 & \color{lightgray}{\XSolidBrush} & \CheckmarkBold & 5830\color{blue}{($\mathbf{3.7}\times$)}  & 77.5(-2.5) \\
    		~ & ~ & 1.1 & \CheckmarkBold & \CheckmarkBold & 5830\color{blue}{($\mathbf{3.7}\times$)} & 78.8(-1.2) \\
    		\cline{2-7}
    		~ & \multirow{4}*{0.5} 
    		& 2.3 & \color{lightgray}{\XSolidBrush} & \color{lightgray}{\XSolidBrush} & 3150($\mathbf{2.0}\times$) & 79.1\color[RGB]{17, 122, 101}{$\mathbf{(-0.9)}$} \\
    		~ & ~ & 2.3& \CheckmarkBold & \color{lightgray}{\XSolidBrush} & 3150($\mathbf{2.0}\times$) & 79.9\color[RGB]{17, 122, 101}{$\mathbf{(-0.1)}$} \\
    		~ & ~ & 2.3& \color{lightgray}{\XSolidBrush} & \CheckmarkBold & 3106($\mathbf{1.9}\times$) & 80.3\color[RGB]{17, 122, 101}{$\mathbf{(+0.3)}$} \\
    		~ & ~ & 2.3& \CheckmarkBold & \CheckmarkBold & 3106($\mathbf{1.9}\times$) & 80.6\color[RGB]{17, 122, 101}{$\mathbf{(+0.6)}$} \\
    		\cline{2-7}
    		~ & 1
    		& 4.6 & \color{lightgray}{\XSolidBrush}  &  \color{lightgray}{\XSolidBrush} & 1597 & 80.0 \\
    		\Xhline{1.0pt}
    	\end{tabular}
    }
    \label{tab:deit}
    \vspace{-0.5cm}
\end{table}

%% file: compare_dvt.tex
\begin{table*}[tp]
	\centering
    \caption{Comparisons between DynamicViT and our SiT on DeiT.    }
    \setlength\tabcolsep{2pt}
    \resizebox{0.95\textwidth}{!}{
    	\begin{tabular}{l|cc|cc|cc}
    		\Xhline{1.0pt}
    		\multirow{3}*{Model} & \multirow{3}*{\makecell[c]{FLOPs \\ ratio}} & \multirow{3}*{\makecell[c]{\#FLOPs \\ (G)}} & \multicolumn{2}{c|}{DynamicViT} & \multicolumn{2}{c}{SiT} \\
    		~ & ~ & ~ & Throughput & ImageNet & Throughput & ImageNet \\
    		~ & ~ & ~ & (image/s) & Top-1(\%) & (image/s) & Top-1(\%) \\
    		\hline
    		\multirow{2}*{DeiT$_{P}$-S} 
    		& \multirow{1}*{0.25} 
    		& 1.1 & 6254($\mathbf{3.8}\times$)  & 65.6(-14.2) & \textbf{6413}\color{blue}{($\mathbf{3.9}\times$)} & \textbf{75.9}\color[RGB]{17, 122, 101}{$\mathbf{(-3.9)}$} \\
    		\cline{2-7}
    		~ & \multirow{1}*{0.5} 
    		& 2.3 & 3248($\mathbf{2.0}\times$) & 78.4(-1.4) & \textbf{3308}\color{blue}{($\mathbf{2.0}\times$)} & \textbf{79.4}\color[RGB]{17, 122, 101}{$\mathbf{(-0.4)}$} \\
    		\cline{2-7}
    		~ & 1
    		& 4.6 & 1637 & 79.8 & 1637 & 79.8 \\
    		\hline
    		\multirow{2}*{DeiT$_{C}$-S}
    		& \multirow{1}*{0.25} 
    		& 1.1 & 5689($\mathbf{3.6}\times$) & 73.4(-6.6) & \textbf{5898}\color{blue}{($\mathbf{3.7}\times$)} & \textbf{78.4}\color[RGB]{17, 122, 101}{$\mathbf{(-1.6)}$} \\
    		\cline{2-7}
    		~ & \multirow{1}*{0.5} 
    		& 2.3 & 3092($\mathbf{1.9}\times$) & 79.2(-0.8) & \textbf{3150}\color{blue}{($\mathbf{2.0}\times$)} & \textbf{79.9}\color[RGB]{17, 122, 101}{$\mathbf{(-0.1)}$} \\
    		\cline{2-7}
    		~ & 1
    		& 4.6 & 1597 & 80.0 & 1597 & 80.0 \\
    		\Xhline{1.0pt}
    	\end{tabular}
    }
    \label{tab:dynamic}
    \vspace{-0.5cm}
\end{table*}

%% file: swin.tex
\begin{table}[h]
	\centering
    \setlength\tabcolsep{8.0pt}
    \resizebox{0.9\linewidth}{!}{
    	\begin{tabular}{c|c|c|c|c}
            \Xhline{1.0pt}
            \multirow{2}*{Model} & \multicolumn{2}{c|}{Baseline} & \multicolumn{2}{c}{Baseline+SiT} \\
            ~ & Throughput & Top-1 & Throughput & Top-1 \\
            \hline
            Swin-T & 1023 & 81.2 & 1183 \color[RGB]{17, 122, 101}{(\textbf{+15.6\%}}) & 81.2 \\
            \hline
            Swin-S & 652 & 83.0 & 855 \color[RGB]{17, 122, 101}{(\textbf{+31.1\%}}) & 83.0 \\
            \Xhline{1.0pt}  
    	\end{tabular}
    }
    \caption{SiT for hierarchical networks.}
    \label{tab:swin}
\end{table}